\documentclass[compsoc]{IEEEtran}
\usepackage{amsmath}
\usepackage{amssymb}
\usepackage{amsfonts}
\usepackage{nicefrac}
\usepackage{graphicx}
\usepackage{lipsum}
\usepackage[dvipsnames]{xcolor}
\usepackage{booktabs}
\usepackage{multirow}
\usepackage{float}
\usepackage{url}
\usepackage{siunitx}
\usepackage[T1]{fontenc}
\usepackage{orcidlink}
\usepackage{outlines}
\usepackage{subcaption}
\usepackage{placeins}
\usepackage{breqn} 

\usepackage{bm}

\title{Biased Heritage: How Datasets Shape Models in Facial Expression Recognition}

\author{
    Iris~Dominguez-Catena\orcidlink{0000-0002-6099-8701}, 
    Daniel~Paternain\orcidlink{0000-0002-5845-887X}, 
    Mikel~Galar\orcidlink{0000-0003-2865-6549}, 
    MaryBeth~Defrance\orcidlink{0000-0002-6570-8857}, 
    Maarten~Buyl\orcidlink{0000-0002-5434-2386}, 
    and~Tijl~De~Bie\orcidlink{0000-0002-2692-7504}%
    \IEEEcompsocitemizethanks{
        \IEEEcompsocthanksitem Iris Dominguez-Catena, Daniel Paternain, and Mikel Galar are with the Institute of Smart Cities (ISC), Public University of Navarre, Pamplona, Spain. E-mail: 
        \{iris.dominguez, daniel.paternain, mikel.galar\}@unavarra.es.
        \IEEEcompsocthanksitem MaryBeth Defrance, Maarten Buyl, and Tijl De Bie are with Ghent University, Ghent, Belgium. E-mail: \{marybeth.defrance,maarten.buyl,tijl.debie\}@ugent.be.}
}

\IEEEtitleabstractindextext{%
\begin{abstract}
    In recent years, the rapid development of artificial intelligence (AI) systems has raised concerns about our ability to ensure their fairness, that is, how to avoid discrimination based on protected characteristics such as gender, race, or age. While algorithmic fairness is well-studied in simple binary classification tasks on tabular data, its application to complex, real-world scenarios—such as Facial Expression Recognition (FER)—remains underexplored. FER presents unique challenges: it is inherently multiclass, and biases emerge across intersecting demographic variables, each potentially comprising multiple protected groups. We present a comprehensive framework to analyze bias propagation from datasets to trained models in image-based FER systems, while introducing new bias metrics specifically designed for multiclass problems with multiple demographic groups. Our methodology studies bias propagation by (1) inducing controlled biases in FER datasets, (2) training models on these biased datasets, and (3) analyzing the correlation between dataset bias metrics and model fairness notions. Our findings reveal that stereotypical biases propagate more strongly to model predictions than representational biases, suggesting that preventing emotion-specific demographic patterns should be prioritized over general demographic balance in FER datasets. Additionally, we observe that biased datasets lead to reduced model accuracy, challenging the assumed fairness-accuracy trade-off.
    
\smallskip
\centering
Code is available at \url{https://github.com/arin-upna/biased-heritage}
\end{abstract}

\begin{IEEEkeywords}
Facial expression recognition, fairness, demographic bias, dataset bias.
\end{IEEEkeywords}}

\begin{document}
\maketitle
\IEEEdisplaynontitleabstractindextext
	
\section{Introduction}\label{sec:intro}

    The accelerated and widespread deployment of artificial intelligence in recent years, largely driven by advances in machine learning models, has raised concerns about the potential harmful effects of this technology~\cite{Adams-Prassl2023}. A prominent concern is the possible discrimination these algorithms may impose on individuals, treating them differently based on protected characteristics such as gender, race, or age~\cite{Curto2024,Bernhardt2022,Chinta2024}. These concerns are particularly relevant in applications like Facial Expression Recognition (FER), which directly interact with users through facial image analysis to infer their emotional states. As FER systems are increasingly deployed in sensitive domains such as healthcare~\cite{Werner2022}, assistive robotics~\cite{Nimmagadda2022}, and public safety~\cite{Mou2023}, ensuring fair and unbiased treatment across different demographic groups becomes crucial. For instance, an unfair automatic pain recognition system could lead to untreated pain in infants from minority groups, directly impacting their healthcare outcomes~\cite{Giordano2024}. Under the umbrella term of algorithmic fairness, numerous studies have attempted to mitigate these potential injustices in general systems~\cite{Jobin2019,Schwartz2022}, but fewer have focused on complex systems like FER. The objective seems clear: designing systems that treat their users fairly and without discrimination.

    Unfortunately, the concept of fairness is neither universal, objective, nor simple to define~\cite{Mitchell2021}. Several formal definitions of fairness coexist~\cite{Mehrabi2021}, often translated into quantifiable metrics of unfairness (bias) to guide model debiasing and comparison~\cite{Hort2023}. While most algorithmic fairness literature focuses on simple binary problems~\cite{Verma2018}, real-world applications like FER present additional challenges: models must classify multiple emotion categories while being fair across several demographic attributes with multiple groups each. Although some fairness definitions have been extended to handle these complex scenarios~\cite{Putzel2022}, their practical implementation as bias metrics requires careful adaptation. In this work, we combine established metrics~\cite{Dominguez-Catena2022,Blakeney2022} for bias in models with novel adaptations to better understand bias in FER systems.

    \begin{figure}[htb]
        \centering
        \begin{subfigure}[b]{\columnwidth}
            \centering
            \includegraphics[width=\columnwidth, trim=0 7.1cm 0 0, clip]{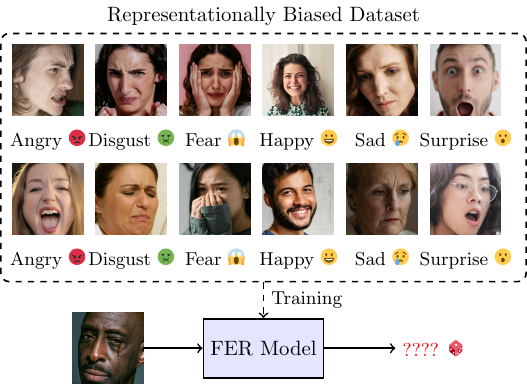}
            \caption{Example of racial representational bias}
            \label{fig:racial_bias}
        \end{subfigure}
        
        \begin{subfigure}[b]{\columnwidth}
            \centering
            \includegraphics[width=\columnwidth, trim=0 0 0 7cm, clip]{images/fig_1.pdf}
            \caption{Example of gender stereotypical bias}
            \label{fig:gender_bias}
        \end{subfigure}
        \caption{Examples of dataset bias affecting model training. (a) shows an example of racial representational bias, where certain demographic groups are overrepresented in the training set. (b) illustrates gender stereotypical bias, where specific emotions are associated with particular demographic groups (e.g., happiness expressions predominantly from female subjects).}
        \label{fig:bias_examples}
    \end{figure}

    While most research on algorithmic bias has focused on measuring bias in model outputs, these metrics alone cannot reveal the origins of the bias~\cite{Adams-Prassl2023}. These sources can be diverse~\cite{Suresh2021}, ranging from data collection methods to the choice of evaluation metrics for the system, but most manifest as alterations in the training or evaluation dataset, called dataset bias. Previous work has analyzed and decomposed dataset bias into two basic and independent subtypes~\cite{Dominguez-Catena2024}: representational bias and stereotypical bias, as illustrated in Fig.\ref{fig:bias_examples}. Representational bias (Fig.\ref{fig:racial_bias}) refers to the overall representation of demographic groups in the dataset, such as the strong Western bias in most image datasets~\cite{Mandal2021}. In contrast, stereotypical bias (Fig.\ref{fig:gender_bias}) focuses on variations in this representation across different classes of the problem, encoding stereotypes about the represented groups. For example, in many FER datasets, women are often overrepresented in the \textit{happy} class while underrepresented in the \textit{angry} class~\cite{Dominguez-Catena2024}. Several metrics exist for both types of selection bias~\cite{Dominguez-Catena2024,Dominguez-Catena2025}, which have already provided insights into the bias properties of different datasets. Nevertheless, there are no comprehensive studies linking dataset bias to model bias as measured by both dataset and model bias metrics.

    The gap between understanding dataset bias and model unfairness is particularly pronounced in complex domains like FER, where multiclass imbalances further complicate the bias assessment. Motivated by this problem, in this work, we focus on studying the propagation of biases from datasets to models and the relationship between bias metrics in datasets and models. By tracing the propagation of bias in the context of FER, we aim to provide insights into both the sources of bias and their downstream effects on model fairness. To do so, we propose a methodology based on systematically inducing bias in a dataset, and then training models to observe the effects of said bias. More specifically, we start with a large FER dataset, AffectNet~\cite{Mollahosseini2019}, and generate biased versions with different types and intensities of bias across two demographic variables. We focus on gender bias and racial bias, comparing how bias manifests differently between a binarily codified variable (gender) and a multigroup codified variable (race)\footnote{We acknowledge that race and gender are social constructs, not discrete biological categories~\cite{Smedley2018}. In this work, we use the terms to refer solely to apparent phenotypic characteristics visible in facial images, as perceived by observers. While this simplifies complex identities~\cite{Morning2011}, it allows us to study potential discriminatory effects in AI systems based on visible traits. The limitations of this approach are further discussed in Section~\ref{ssec:dataset}.}. We then train a ResNet50 model~\cite{He2015} on each of the resulting biased datasets and analyze the models' predictions for potential traces of bias. We then perform a detailed analysis of bias propagation patterns, investigating how different dataset bias metrics in our intentionally biased datasets correlate with and predict various forms of model unfairness.

    Some of our main findings are:
    \begin{itemize}
        \item Different propagation patterns emerge for representational and stereotypical biases. While current model bias metrics show higher sensitivity to stereotypical bias effects, this could mask important but currently undetectable impacts of representational bias.
        \item Differences manifest not only between dataset bias types, but also in how they are manifested. For example, stereotypical bias shows stronger propagation when affecting multiple emotion classes simultaneously\textemdash an effect that current dataset bias metrics fail to capture.
        \item Surprisingly, our results reveal that dataset bias significantly degrades both model fairness and accuracy, strongly challenging the widely accepted fairness-accuracy trade-off assumption~\cite{Menon2018} in FER. This suggests that bias mitigation at the dataset level could simultaneously improve both model fairness and performance.
    \end{itemize}
    
    In summary, the main contributions of this work are:
    
    \begin{outline}
        \1 We systematically analyze bias propagation from datasets to models in FER, revealing distinct impact patterns for different types of dataset bias and establishing their relationship with both model bias metrics and model performance.
        \1 We propose new model bias metrics that adapt classical fairness constraints to handle both multiple classes and demographic groups, extending concepts like \textit{equalized odds}, \textit{equal opportunity} and \textit{demographic parity} to the FER context.
        \1 We demonstrate that in FER, careful dataset design focusing on preventing emotion-specific demographic patterns may be more crucial than achieving perfect demographic balance.
    \end{outline}

    This work bridges the parallel developments in dataset bias and model bias, illustrating the ultimate impact of dataset bias and offering new insights into the origins of various biases within models. By mapping the bias propagation throughout the machine learning pipeline, this research enables improved bias mitigation in AI systems and facilitates the prediction of biases based on the bias characteristics of the training datasets.

\section{Background and related work}\label{sec:related}

    While fairness is typically focused on the output of the models, unfairness or bias can originate from different phases in the ML pipeline, such as the data collection process, annotation procedures, and the demographics of the source population~\cite{Suresh2021}. Many such biases are reflected in the datasets used to train the models. In this work, we distinguish between model bias, making reference to the bias in the model predictions, and dataset bias~\cite{Dominguez-Catena2024}, referring to the systematic inaccuracies or patterns in demographic representation present in the data used to train machine learning models, and which aggregates biases from previous sources. Regardless of the source of the bias, if it is present in the dataset, it has the potential to affect the model's performance and predictions. 
    
    Based on this classification of bias, in this section we cover some notions and previous works that contextualize our research. First, Section~\ref{ssec:rel_model} presents some notions about algorithmic fairness and its relationship with bias measurement in models. Then, Section~\ref{ssec:rel_dataset} complements them by focusing on dataset bias, the origin of many biases. Finally, Section~\ref{ssec:rel_fer} deals with bias in the context of FER.

	\subsection{Fairness and model bias}\label{ssec:rel_model}

    The deployment of artificial intelligence models in recent years has led to a growing need to understand the potential negative aspects of this technology. One concerning aspect is the possibility that these systems may treat people unfairly based on certain protected characteristics, such as sex, gender, race, or age, among others~\cite{Mehrabi2021}. This problem has become particularly relevant with the leap of artificial intelligence from the industrial sphere to applications in more direct contact with society, including the medical field, politics, and advertising~\cite{Dwivedi2021}.

    To monitor and address this issue, multiple mathematical definitions of algorithmic fairness have been designed, which can be broadly categorized as group fairness, individual fairness, or causal fairness~\cite{Verma2018}. Group fairness definitions ensure equal outcomes or treatment across demographic groups, such as equal error rates or selection rates~\cite{Hardt2016,Zafar2017}. In contrast, individual fairness definitions focus on providing consistent and coherent treatment of similar individuals, independent of group membership~\cite{Dwork2012}. Causal fairness definitions take a more nuanced approach, using causal reasoning to identify and mitigate unintended influences of protected attributes on algorithmic decisions~\cite{Kusner2018}. Additionally, there are other mixed approaches that combine the properties of these categories~\cite{Kearns2018}. From these approaches, group fairness is the most common, as it aligns well with established legal and policy frameworks that regulate discrimination while providing a clear explanation of the properties of a system~\cite{Corbett-Davies2018}.

    Among the different possible definitions of fairness within the framework of group fairness, the most popular are \textit{equalized odds} and its relaxed form \textit{equal opportunity}~\cite{Hardt2016}, and \textit{demographic parity}~\cite{Zafar2017}, although both the names and exact definitions vary considerably between authors~\cite{Verma2018}. In general, the definitions are based on statistical requirements on the predictions of a model, demanding certain properties to be equal across different demographic groups, such as the probability of predicting the positive class (\textit{demographic parity}). However, most of the bias metrics derived from these fairness definitions are specific to binary classification problems and only consider two demographic groups (a general population and a protected group). Recent work has begun to extend these definitions to multiclass problems~\cite{Denis2023,Blakeney2022,Putzel2022,Dominguez-Catena2022} and multiple demographic groups~\cite{Blakeney2022,Putzel2022,Dominguez-Catena2022}, as well as continuous demographic variables~\cite{Jiang2022}.

    To make these definitions operational, many of them have been converted into measures of bias or unfairness~\cite{Mehrabi2021}. Rather than requiring strict equality between groups, these measures quantify the deviation from ideal fairness in model predictions. Measurement of fairness through these metrics is more practical for real-world applications, where minor deviations may be acceptable and do not indicate systematic bias. This approach also helps evaluate and guide bias mitigation strategies~\cite{Hort2023}. Section~\ref{ssec:measurement_model} provides a detailed analysis of the metrics relevant to our context.

    It is important to note that different fairness definitions, while based on common values and intuitions, are not equivalent and can even be mathematically incompatible~\cite{Miconi2017}. When these definitions are used as optimization constraints for bias mitigation, it becomes impossible to minimize all metrics simultaneously. However, when evaluating models that have not been optimized to reduce bias, these metrics can align in their assessment, as our results demonstrate.

    \subsection{Dataset bias}\label{ssec:rel_dataset}

    The biases in datasets can manifest in different forms, from biases in the data samples themselves (framing bias) to biases in the general distribution of examples across different demographics (selection bias)~\cite{Fabbrizzi2022}. In this work, we focus on selection bias, as it allows for precise quantitative analysis while having a significant impact on the trained models, similar to class imbalance situations~\cite{Fernandez2018}. As previously illustrated in Figure~\ref{fig:bias_examples}, prior work~\cite{Dominguez-Catena2024} has differentiated two main families of selection bias:
    
    \begin{outline}
        \1 \textbf{Representational bias} refers to the global presence or absence of certain demographic groups in a dataset. For example, the tendency of a dataset to include only subjects within a certain age range, like the prevalence of people in their twenties in FER datasets~\cite{Dominguez-Catena2025}. This type of bias can also be split into several specific subtypes, such as \textit{richness} (number of groups represented), \textit{evenness} (homogeneity of the group representations) and \textit{dominance} (prevalence of the most represented group over the rest). Despite this, it has been suggested that combining a general representational bias measure with an evenness measure can be sufficient to characterize representational bias, as evenness and dominance correlate strongly~\cite{Dominguez-Catena2024}.
        \1 \textbf{Stereotypical bias}, on the other hand, occurs in datasets with different demographic compositions in each class. For instance, a FER dataset with more women in the \textit{happy} class and more men in the \textit{angry} class is stereotypically biased, as happens in AffectNet and other FER datasets~\cite{Dominguez-Catena2024}.
    \end{outline}

    In Section~\ref{ssec:measurement_dataset} we review the relevant metrics employed to quantify these two types of dataset bias.
    
	\subsection{Facial Expression Recognition (FER)}\label{ssec:rel_fer}

    Facial Expression Recognition is one of the most prominent problems in Affective Computing~\cite{Assuncao2022}, providing systems with a simple way to observe and identify user emotions through facial images. This approach offers two primary benefits: low hardware requirements and the possibility of training models using readily available data from the Internet, such as stock pictures. FER systems have diverse applications, including assistive robotics~\cite{Nimmagadda2022}, healthcare~\cite{Werner2022}, public safety~\cite{Mou2023}, and general human-computer interaction tasks~\cite{Bakariya2023}. In these applications, FER systems often interact directly with end-users, many of whom are non-experts and possibly unaware of the AI-driven nature of the technology.
    
    Image classification is the predominant approach to FER, involving the classification of a single facial image into a discrete set of emotional states. The most widely used emotion set consists of Ekman's six basic emotions~\cite{Ekman1971} (\textit{anger}, \textit{disgust}, \textit{fear}, \textit{sadness}, \textit{surprise}, and \textit{happiness}), commonly including a \textit{neutral} state. Alternative codifications exist, such as continuous valence-arousal models~\cite{Kollias2021}, or emotion sets for specialized contexts, such as pain detection~\cite{Werner2022}. Despite these alternatives, the classification system based on Ekman's model remains the most common due to its broad acceptance and the relative simplicity of manual annotation in FER datasets~\cite{Goodfellow2013,Zhang2017,Mollahosseini2019}.
    
    FER is usually approached with deep learning techniques, such as residual neural networks (ResNet)~\cite{He2015}, as they provide good performance and robustness. However, training these networks requires significant computational power and large amounts of data. As a result, the latest FER datasets predominantly use data sourced from the Internet~\cite{Nidhi2023}, not only for the quantity available but also for its general diversity.
    
    Unfortunately, FER datasets are vulnerable to bias issues. While the multiclass nature of FER complicates research, coupled with the large number of potentially impacted demographic groups, internet-sourced FER datasets have demonstrated both representational and stereotypical biases~\cite{Dominguez-Catena2024}. Notably, stereotypical bias manifests in larger quantities compared to traditional datasets collected in controlled laboratory settings. Although most FER datasets lack explicit demographic information about depicted subjects, such data is correlated with their visual appearance, enabling the identification of characteristics such as age, gender, and race. Consequently, models can indirectly discriminate based on these factors. Moreover, facial images can reveal additional social attributes, such as social class~\cite{Bjornsdottir2017}, potentially worsening bias dynamics within FER datasets.

    Recent research has highlighted biases concerning gender, race, and age in FER models~\cite{Kim2021,Ahmad2022,Xu2020,Domnich2021,Jannat2021,Deuschel2021,Poyiadzi2021,Raina2022}. These studies include evaluations of age-related biases in commercial systems~\cite{Kim2021} and comprehensive analyses of age, gender, and race biases using specialized video databases~\cite{Ahmad2022}, while approaches using synthetic faces have helped isolate and demonstrate racial biases~\cite{Raina2022}. Research models trained on Internet-collected data have been evaluated for age, gender, and race biases using datasets with known demographic information~\cite{Xu2020}. Several studies have focused on specific bias types, such as gender bias~\cite{Domnich2021,Jannat2021}, or the relationship between facial action unit prediction and demographic attributes like gender and skin color in established datasets~\cite{Deuschel2021}. Age bias has also been examined in Internet-sourced datasets~\cite{Poyiadzi2021}. Following these bias discoveries, various mitigation strategies have been proposed~\cite{Xu2020,Jannat2021,Poyiadzi2021}. Additionally, guidelines have been developed to assess and reduce potential risks in FER technology deployment~\cite{Hernandez2021}.

\section{Bias metrics}\label{sec:measurement}

    In this section, we present the metrics used to quantify bias in both datasets and trained models. These metrics serve different but complementary purposes: dataset bias metrics measure systematic imbalances in training data, while model bias metrics evaluate fairness in model predictions. For dataset bias metrics (Section~\ref{ssec:measurement_dataset}), we employ both classical statistical measures and similarity-based metrics to quantify representational and stereotypical biases. For model bias metrics (Section~\ref{ssec:measurement_model}), we combine established metrics applicable to multiclass and multi-group scenarios with new adaptations of classical fairness constraints.

    To ensure clear notation throughout this section, we first establish some common conventions used across both dataset and model bias metrics:

    \begin{itemize}
        \item We define $\mathcal{A}$ as the set of demographic groups defined by the value of a protected attribute. For example, if $\mathcal{A}$ stands for \textit{gender}, we may approximate it with the set $\mathcal{A} = \{\text{Male}, \text{Female}\}$.
        \item We define $\mathcal{C}$ as the set of classes of the problem. In the case of FER, these are expressions.
        \item We define $X$ as a population of $n = |X|$ samples.
        \item We define $n^a$ with $a \in \mathcal{A}$ as the number of samples in $X$ where the subject belongs to the demographic group $a$. Hence $\sum_{a \in \mathcal{A}} n^a = n$.
        \item Similarly, we define $n_y$ with $y \in \mathcal{C}$ as the number of samples labeled as class $y$.
        \item Similarly, we define $p^a$ and $p_y$ as the proportion of samples from the population $X$ that belong to a certain group $p^a = \frac{n^a}{n}$, or are labeled with a certain class $p_y = \frac{n_y}{n}\;.$
    \end{itemize}

\subsection{Dataset bias metrics}\label{ssec:measurement_dataset}

    In this work, we use a previously proposed set of dataset bias metrics composed of three metrics~\cite{Dominguez-Catena2024}, and we complement these metrics with another set based on dataset similarity~\cite{Dominguez-Catena2025}. The total number of metrics is six, four for representational bias and two for stereotypical bias.

    \subsubsection{Representational bias metrics}

    Following recommendations from previous studies~\cite{Dominguez-Catena2024}, we employ two distinct classical metrics for representational bias: one for general representational bias and another for evenness.

    \vspace{0.3em}\noindent\textbf{Effective Number of Species (ENS)}~\cite{Jost2006}. This representational bias metric robustly adapts the number of demographic groups represented in the dataset. It employs the notion of the \textit{effective number of groups}, which uses the Shannon entropy of the group proportions. Intuitively, it computes a weighted count over the number of groups in the dataset where the weight is only 1 if the group's representation is $\frac{1}{|\mathcal{A}|}$ and lower the more it is under-or over-represented (such that the group's weight is lower bounded by 0). This metric offers a similar idea to the number of represented groups, but varies less drastically when introducing small-sized groups. It has a lower bound of $1$, for a dataset with only one group represented, and an upper bound of $|\mathcal{A}|$, the number of demographic groups.
    
    \begin{equation}
        \text{ENS}(X) = \exp\left({-\sum_{a\in \mathcal{A}}{p^a \ln p^a}}\right)\;.
    \end{equation}

    In its canonical formulation, ENS measures diversity, the opposite of bias. In this work, we redefine the metric to directly quantify bias by subtracting from its maximum: $|\mathcal{A}| - \text{ENS}$.

    \vspace{0.3em}\noindent\textbf{Shannon Evenness Index (SEI)}~\cite{Pielou1966}. Also based on the Shannon entropy, SEI is a representational bias metric specialized in evenness, one of the main subtypes of this type of bias. It measures the homogeneity among the sizes of different demographic groups, regardless of their total number. The metric ranges from $0$ to $1$, approaching $0$ for the more uneven datasets, and $1$ indicating identical representation among the groups.

    \begin{equation}
        \text{SEI}(X) = \frac{-\sum_{a\in \mathcal{A}} p^a \ln(p^a)} {\ln(|\{a\in \mathcal{A}|n^a > 0\}|)}\;.
    \end{equation}

    Similarly to ENS, we also redefine SEI to directly measure unevenness or bias: $1 - \text{SEI}$. Note that \text{SEI}($X$) can be considered as the \text{ENS}($X$) normalized based on the number of groups in the dataset.

    \vspace{0.6em}\noindent Whereas the previously presented classical metrics describe the statistical properties of a dataset's demographic composition, similarity-based metrics measure the demographic resemblance between the dataset and an ideal unbiased dataset used as a reference point~\cite{Dominguez-Catena2025}. Specifically, we employ the family of bias metrics based on DSAP~\cite{Dominguez-Catena2025}, a methodology for demographic comparison of datasets that uses demographic profiles generated with an auxiliary model. The major advantage of these metrics is their ease of use in comparisons across different types of bias, since the same similarity metric, the Renkonen similarity index~\cite{Renkonen1938}, serves as a base for all the metrics:

    \begin{equation}\label{eq:ds}
        \text{R}(X, X') = 1 - 0.5\sum_{a\in \mathcal{A}}\left|p^a - {p'}^a\right|\;
    \end{equation}
where $p^a$ and ${p'}^a$ refer to demographic proportions over datasets $X$ and $X'$ respectively.

    This index R is bounded between $0$ for completely different demographic profiles and $1$ for identical ones. Using this common definition, the following metrics are derived for representational bias:

    \vspace{0.3em}\noindent\textbf{DSAP-based representational bias ($\text{DS}_R$)}~\cite{Dominguez-Catena2025}. Representational bias is measured by comparing $X$, the dataset of interest, with an unbiased version of the dataset $X^{\text{rep}}$ that represents all demographic groups equally. Formally,

    \begin{equation}
        \text{DS}_R(X) = \text{R}(X, X^{\text{rep}})\;,
    \end{equation}
    where $X^{\text{rep}}$ has demographic proportions $p^{\text{rep},a}$ with
    \begin{equation}
        p^{\text{rep},a} = \frac{1}{|\mathcal{A}|}\;.
    \end{equation}

    Since $\text{DS}_R$ measures the similarity to an ideal, we redefine it by subtracting from 1 to directly measure bias: $1-\text{DS}_R$.
    
    \vspace{0.3em}\noindent\textbf{DSAP-based evenness ($\text{DS}_E$)}~\cite{Dominguez-Catena2025}. Evenness is measured by computing $DS$ between the dataset of interest and a dataset representing the same groups as the dataset of interest, but equally.

    \begin{equation}
        \text{DS}_E(X) = \text{R}(X, X^{\text{even}})\;,
    \end{equation}
    where $X^{\text{even}}$ has demographic proportions $p^{\text{even},a}$ with
    \begin{equation}
        p^{\text{even},a} = \begin{cases}
            \frac{1}{|\left\{a\in \mathcal{A}|p^a>0\right\}|} & \text{if }p^a > 0\\
            0 & \text{otherwise}\;.\\
        \end{cases}
    \end{equation}

    Likewise $\text{DS}_R$, we redefine this measure to obtain a measure of bias ($1-\text{DS}_E$).

    \subsubsection{Stereotypical bias metrics}

    For stereotypical bias, we employ one of the classical metrics, Cramer's V~\cite{Dominguez-Catena2024}.

    \vspace{0.3em}\noindent\textbf{Cramer's V (V)}~\cite{Cramer1991}. Cramer's V is a measure of stereotypical bias based on the correlation between the target variable and a demographic variable. It is bounded in the unitary range, in this case with $0$ corresponding to unbiased datasets and $1$ to stereotypically biased ones. It is defined as follows:

    \begin{equation}
        \text{V}(X) = \sqrt{ \frac{\chi^2(X)/n}{\min(|\mathcal{A}|-1,|\mathcal{C}|-1)}}\;,
    \end{equation}
    where $\chi^2(X)$ is the Pearson's ${\chi^2}$ statistic, defined as:
    \begin{equation}
        \chi^2(X)=\sum_{a \in \mathcal{A}}\sum_{y \in \mathcal{C}}\frac{(n^a_y-\frac{n^a n_y}{n})^2}{\frac{n^a n_y}{n}}\;.
    \end{equation}

    \vspace{0.6em}\noindent Similarly to representational bias, we also employ a metric that measures stereotypical bias through demographic dataset comparison~\cite{Dominguez-Catena2025}:

    \vspace{0.3em}\noindent\textbf{DSAP-based stereotypical bias ($\text{DS}_S$)}~\cite{Dominguez-Catena2025}. Stereotypical bias is measured by computing $DS$ between the population of each class in the dataset of interest and the rest of the dataset, then averaging across classes.

    \begin{equation}
        \text{DS}_S(X) = \frac{1}{|\mathcal{C}|}\sum_{y\in \mathcal{C}}\text{R}(X_y, X_{\bar{y}})\;,
    \end{equation}
    where $X_y$ refers to the subset of the dataset belonging to the class $y$, and $X_{\bar{y}}$ to the rest of the dataset. As before, to directly measure bias, we consider $1-\text{DS}_S$.

\subsection{Model bias metrics}\label{ssec:measurement_model}

    In this section, we focus on measuring biases in trained models by reviewing and extending existing model bias metrics from the literature. The context of FER requires metrics that support multiclass problems, as well as demographic variables codified with multiple groups, limiting the applicability of many classic metrics~\cite{Mehrabi2021} like \textit{equalized odds}~\cite{Hardt2016}, \textit{equal opportunity}~\cite{Hardt2016}, and \textit{demographic parity}~\cite{Zafar2017}. Prior work has extended these metrics to multiclass and multigroup settings, but primarily by composing them as a set of fairness constraints rather than as quantitative bias metrics. To address this gap, in Section~\ref{sssec:novel_metrics} we propose a natural adaptation of these notions into measurable metrics. Later in Section~\ref{sssec:preext_metrics}, we discuss several existing metrics that are also directly applicable to our setting, including Overall Disparity (OD)~\cite{Dominguez-Catena2022}, Combined Error Variance (CVE), and Symmetric Distance Error (SDE)~\cite{Blakeney2022}. This completes our final set of 7 model bias metrics.

    Before presenting the metrics, it is necessary to define some common rates used in model evaluation. These rates are based on the model's predictions, which are captured in a raw confusion matrix $\widehat{W}^a$ for each demographic group, where each element $\widehat{W}_{y,\hat{y}}^a$ represents the count of instances with true class $y \in \mathcal{C}$ that the model predicts as class $\hat{y} \in \mathcal{C}$. To simplify subsequent definitions, we normalize this raw matrix as follows:

    \begin{equation}
        W^{a}_{y\hat{y}} = \frac{\widehat{W}^{a}_{y\hat{y}}}{\sum_{\hat{z}\in\mathcal{C}}\widehat{W}^{a}_{y\hat{z}}}.
    \end{equation}

    From this normalized confusion matrix, we can approximate $P(\hat{Y}=\hat{y}|Y=y,A=a)\approx W^{a}_{y\hat{y}}$. This approximation allows us to compute the following group conditional rates.

    \vspace{0.5em}\noindent\textit{True Positive Rate (TPR)}. The TPR is the rate at which the predictions are correct for class $y$ and group $a$:
    \begin{equation}
    \text{TPR}_{y}^a = P(\hat{Y}=y|Y=y,A=a)\;,
    \end{equation}
    which are the diagonal values on the normalized confusion matrix $\text{TPR}_{y}^a = W^a_{yy}$.

    \vspace{0.5em}\noindent\textit{False Negative Rate (FNR)}. The FNR is the rate at which the predictions are incorrect for the class $y$ and protected attribute value $a$:
    \begin{equation}
    \text{FNR}_{y}^a = P(\hat{Y}\neq y|Y= y,A=a)\;,
    \end{equation}
    which can be calculated as $\text{FNR}_{y}^a = 1 - \text{TPR}_{y}^a$.
    
    \vspace{0.5em}\noindent\textit{False Positive Rate (FPR)}. The FPR is the rate at which the class $y$ is (wrongfully) predicted when the true label is \textit{not} $y$:
    \begin{equation}
    \text{FPR}_{y}^a = P(\hat{Y}=y|Y\ne y,A=a)\;,
    \end{equation}
    which can be calculated as
    \begin{equation}\text{FPR}_{y}^a = \frac{1}{\sum_{\substack{z,\hat{z}\in\mathcal{C}\\z\ne\hat{y}}}{W}_{z\hat{z}}^{a}}\sum_{\substack{z\in\mathcal{C}\\z\ne \hat{y}}}{W}_{z\hat{y}}^{a}\;.\end{equation}

    \vspace{0.5em}\noindent\textit{False Detection Rate (FDR)}. The FDR is the rate at which any error (i.e. $Y\ne \hat{Y}$) for group $a$ is predicted to be of class $y$, regardless of the actual class:
    \begin{equation}
    \text{FDR}_{y}^a = P(\hat{Y}=y|Y\ne \hat{Y},A=a)\;,
    \end{equation}
    which differs from the FPR in its normalization (summing over all errors instead of only over errors for $y$). It can thus be calculated as
    \begin{equation}\text{FDR}_{y}^a = \frac{1}{\sum_{\substack{z,\hat{z}\in\mathcal{C}\\z\ne\hat{z}}}{W}_{z\hat{z}}^{a}}\sum_{\substack{z\in\mathcal{C}\\z\ne \hat{y}}}{W}_{z\hat{y}}^{a}\;.
    \end{equation}

    \subsubsection{Novel adaptations}\label{sssec:novel_metrics}

    Based on these auxiliary definitions, we propose a set of model bias metrics that extend classical algorithmic fairness notions \textemdash \textit{equalized odds}~\cite{Hardt2016}, \textit{equal opportunity}~\cite{Hardt2016}, and \textit{demographic parity}~\cite{Zafar2017} \textemdash to multiclass and multigroup problems. Starting from pre-existing formulations adapted to multiple classes and demographic groups~\cite{Putzel2022}, we adapt them as bias metrics. For each fairness constraint, we measure bias by taking the maximum difference between demographic groups and then the average across classes. Using the maximum difference across demographic groups highlights any type of bias situations where certain groups under or overperform, and the average across classes considers that all of them contribute equally to the model's fairness. In addition, these metrics reduce to their classical formulations when $|\mathcal{C}|=2$ and $|\mathcal{A}|=2$.
    
    \vspace{0.5em}\noindent\textbf{Term-by-term Multiclass Equalized Odds (TTEqOdds)}. This is the most restrictive notion of fairness possible in this context, requiring that the group conditional confusion matrices be identical across all groups. We define the bias metric as follows:

    \begin{equation}
        \text{TTEqOdds} = \frac{1}{|\mathcal{C}|^2}\sum_{y\in\mathcal{C}}\sum_{\hat{y}\in\mathcal{C}}\max_{a_1,a_2\in \mathcal{A}}\lvert\text{W}_{y\hat{y}}^{a_1}-\text{W}_{y\hat{y}}^{a_2}\rvert.
    \end{equation}

    \vspace{0.5em}\noindent\textbf{Classwise Multiclass Equalized Odds (CEqOdds)}. This fairness constraint relaxes the TTEqOdds conditions, requiring only that the true positive rates (diagonal of the confusion matrices) are equal across groups, and additionally, that the FDR for each class is the same for all groups. Recall that the FDR is the probability of a certain class being predicted when a mistake occurs. We propose the following definition:

    \begin{equation}
    \begin{aligned}
        \text{CEqOdds} = &\frac{1}{2 |\mathcal{C}|}\sum_{y\in\mathcal{C}}\max_{a_1,a_2\in \mathcal{A}}\lvert\text{TPR}_{y}^{a_1}-\text{TPR}_{y}^{a_2}\rvert + \\
            & \frac{1}{2|\mathcal{C}|}\sum_{\hat{y}\in\mathcal{C}}\max_{a_1,a_2\in \mathcal{A}}\lvert\text{FDR}_{\hat{y}}^{a_1}-\text{FDR}_{\hat{y}}^{a_2}\rvert\;,
    \end{aligned}
    \end{equation}

    \vspace{0.5em}\noindent\textbf{Multiclass Equality of Opportunity (EqOpp)}. In this case, only the TPRs (diagonal of the confusion matrices) are required to be equal across groups.
    
    \begin{equation}
    \begin{aligned}
        \text{CEqOpp} = &\frac{1}{|\mathcal{C}|}\sum_{y\in\mathcal{C}}\max_{a_1,a_2\in \mathcal{A}}\lvert\text{TPR}_{y}^{a_1}-\text{TPR}_{y}^{a_2}\rvert\;.
    \end{aligned}
    \end{equation}

    \vspace{0.5em}\noindent\textbf{Multiclass Demographic Parity (DemPar)}. Demographic parity differs from other fairness constraints as it does not consider the true classes, but only the predictions. It requires that, for each class, the probability of predicting that class, conditional on the demographic group, is equal across groups.

    
    \begin{equation}
    \begin{aligned}
        \text{DemPar} = \frac{1}{|\mathcal{C}|}\sum_{\hat{y}\in\mathcal{C}}\max_{a_1,a_2\in \mathcal{A}}\left|\sum_{y\in\mathcal{C}}{W}_{y\hat{y}}^{a_1}-\sum_{y\in\mathcal{C}}{W}_{y\hat{y}}^{a_2}\right|\;.
    \end{aligned}
    \end{equation}

    \subsubsection{Preexisting metrics}\label{sssec:preext_metrics}

    Aside from these new proposals, we also employ the following pre-existing model bias metrics, which can operate on multiclass problems and multigroup demographic variables.

    \vspace{0.5em}\noindent\textbf{Overall disparity (OD)}~\cite{Dominguez-Catena2022}. This metric focuses on the differences in class TPRs between the different groups. A particularity of this metric is that it does not focus on the absolute differences, but rather on the relative difference of each group compared to the best-recognized group. It is computed in two steps: first, obtaining the intra-class disparity (ID), and then computing the OD by aggregating the ID of all classes:

    \begin{equation}
        \text{ID}_y = \frac{1}{|\mathcal{A}|-1}\sum_{a \in \mathcal{A}} \left(
                {1-\frac{\displaystyle \text{TPR}_y^a}{\displaystyle \max_{a'\in \mathcal{A}}{\text{TPR}_y^{a'}}}}\right),
    \end{equation}
    
    \begin{equation}
        \text{OD} = \frac{1}{|\mathcal{C}|}\sum_{y \in \mathcal{C}}{\text{ID}_y}\;.
    \end{equation}

    \vspace{0.5em}\noindent\textbf{Combined Error Variance (CVE)}~\cite{Blakeney2022}. CVE is originally designed to compare the predictions of two models, analyzing the tendency of some biased models to sacrifice performance in some classes in favor of others. In particular, it analyzes deviations of the false positive and false negative rates (FPR and FNR). This same definition can also be applied to measure bias within a single model by comparing its behavior for a protected demographic group against its behavior on the entire dataset.

    Let $\Delta\text{FPR}^a_i$ denote the relative difference in FPR for class $y$ and protected attribute value $a$ compared to the overall FPR:
    \[
    \Delta\text{FPR}^a_y = \frac{\text{FPR}_y^a - \text{FPR}_y}{\text{FPR}_y}\;,
    \]
    where FPR$_y$ refers to FPR over the whole dataset. The mean relative difference $\overline{\Delta\text{FPR}^a}$ is:
    \[
    \overline{\Delta\text{FPR}^a} = \frac{1}{|\mathcal{C}|}\sum_{y\in\mathcal{C}}\Delta\text{FPR}^a_y\;.
    \]
    Analogous definitions apply for FNR, yielding $\Delta\text{FNR}^a_y$ and $\overline{\Delta\text{FNR}^a}$.
    
    Using the Euclidean distance as distance function $d$, the Combined Error Variance is defined as:
    \begin{equation}
    \begin{aligned}
        \text{CEV} = \frac{1}{|\mathcal{C}||\mathcal{A}|}\sum_{y\in\mathcal{C}}\sum_{a\in \mathcal{A}}d( & (\overline{\Delta\text{FPR}^a},
        \overline{\Delta\text{FNR}^a}),\\ 
         & (\Delta\text{FPR}^a_y,
        \Delta\text{FNR}^a_y))^2\;.
    \end{aligned}
    \end{equation}
    
    \vspace{0.5em}\noindent\textbf{Symmetric Distance Error (SDE)}~\cite{Blakeney2022}. SDE is similar to CVE, but focuses on the tendency of biased models to change the mode of failure, that is, the balance between FPR and FNR for a given class. As with CVE, it can be easily applied for model bias measurement. 

    \begin{equation}        
        \text{SDE} = \frac{1}{|\mathcal{C}||\mathcal{A}|}\sum_{y\in\mathcal{C}}\sum_{a\in \mathcal{A}}|\Delta\text{FPR}^a_y -
        \Delta\text{FNR}^a_y|\;.
    \end{equation}

\section{Methodology}\label{sec:methodology}

\begin{figure*}
        \centering
        \includegraphics[width=0.85\textwidth]{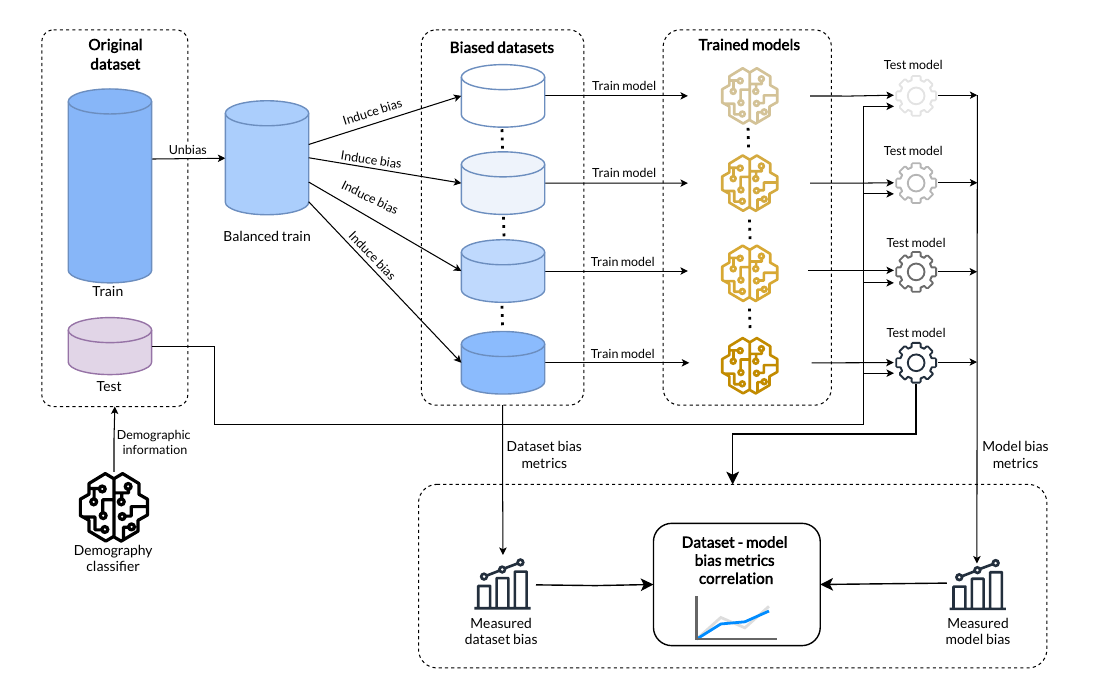}
        \caption{Summary of the methodology}
        \label{fig:methodology}
    \end{figure*}

    In this section, we present our proposed methodology for measuring bias propagation from the training dataset to the trained model. A summary of this methodology is provided in Figure~\ref{fig:methodology}. We begin with a demographically labeled dataset or, if demographic labels are unavailable, use an auxiliary model to estimate them. This information is then used to create a balanced subset of the dataset with respect to a demographic component. This balanced subset serves as a base for further subsampling, where datasets with a predetermined amount of induced bias are generated. The induced bias is measured using the dataset metrics discussed in Section~\ref{ssec:measurement_dataset}. Subsequently, we train a model on each biased dataset and, using the metrics outlined in Section~\ref{ssec:measurement_model}, quantify the bias in the model's predictions. By analyzing the correlations between the induced dataset bias and the resulting model bias, we can identify the propagation mechanisms of different types of biases.

    The rest of the section is organized as follows. First, Section~\ref{ssec:dataset} introduces the chosen base dataset, and then Section~\ref{ssec:induction} explains the bias induction process applied to it. After that, Section~\ref{ssec:overview} provides an overview of the experiments considered. Finally, Section~\ref{ssec:setup} details the experimental setup on which we ran the experiments.
	
	\subsection{Source dataset}\label{ssec:dataset}

    Over the years, a considerable number of datasets have been created for FER~\cite{Nidhi2023}. Following the trend towards in-the-wild datasets from Internet sources, we chose AffectNet~\cite{Mollahosseini2019} for this work. AffectNet is one of the largest public datasets for FER, composed of $420\,299$ images with a native resolution of $425\times425$ pixels, sourced from Internet searches. The images in the dataset are labeled in two different schemes: a continuous valence-arousal emotion codification, and a discrete codification based on Ekman's six basic emotions~\cite{Ekman1971}. For this work, we focus on the subset labeled with Ekman's six basic emotions~\cite{Ekman1971} ($286\,341$ images), since most model bias metrics are designed for classification models. We use the provided train and test partitions, with $282\,853$ and $3\,488$ images, respectively. While the test partition is not demographically balanced, this does not compromise our analysis since the employed model bias metrics inherently account for and normalize demographic differences. Moreover, attempting to balance this partition through subsampling would further reduce its already limited size, potentially degrading the reliability of our measurements.

    Unfortunately, any analysis of dataset biases is limited by the availability of sufficient demographic information. This is especially common among in-the-wild datasets such as those used for FER, like AffectNet. To overcome this, we used the FairFace model~\cite{Karkkainen2021} to infer apparent gender and race from each image, following previous works~\cite{Yang2022,Ghosh2022,Dominguez-Catena2024}. Although the model additionally provides an age prediction, for our experiments we focus only on gender and race, as examples of binary and non-binary codified demographic variables. These two variables are chosen because they have been extensively studied in the context of bias, both within and outside FER~\cite{Garrido-Munoz2023,Sham2022,Dominguez-Catena2023,Domnich2021}. To maximize the robustness of these approximated labels, we apply to AffectNet the same preprocessing that was used for FairFace. In particular, we preprocess images using the Max-Margin Object-Detection CNN face extractor~\cite{King2015} before obtaining the predicted demography with FairFace.

    It is important to consider the inherent limitations of this approach. FairFace offers very high accuracy across several datasets, above $75.4\%$ in race prediction and $92.0\%$ in gender prediction. However, the resulting demographic labels are just a proxy for the true demography of the people represented in the dataset, which would ideally be self-reported. Additionally, our analysis is subject to the constraints of this model, which reports gender in a binary way (labeled as Female -- F or Male -- M) and race as one of a discrete set of seven racial groups (Black -- B, East Asian -- EA, Indian -- I, Latino-Hispanic -- LH, Middle Eastern -- ME, Southeast Asian -- SA, and White -- W). However, as our focus is studying group-level trends, we consider these approximate labels a reasonable substitute.

    The demographic profile of AffectNet, as estimated by FairFace, is shown in Table~\ref{tab:cont_affectnet}. For each emotion, the dominant gender and race is shown in bold. Overall, the dataset is balanced by gender, but some clear disparities can be observed in the independent emotions, indicating the presence of stereotypical bias. Race-wise, the dataset is clearly dominated by the White category, showing a strong representational bias.

    \begin{table*}[tb]
		\centering
        \sisetup{
            table-alignment-mode = format,
            table-number-alignment = right,
            table-text-alignment = right,
            table-format=5.0,
            group-minimum-digits = 4,
            detect-weight=math,
            mode=text
            }
        \caption{Original contingency table for the train partition of AffectNet.}\label{tab:cont_affectnet}
		\begin{tabular}{lSSSSSSSSS}
			\toprule
            \multirow{2}{*}{\textbf{Class}} & \multicolumn{2}{c}{\textbf{Gender}}  &  \multicolumn{7}{c}{\textbf{Race}} \\
            \cmidrule(lr){2-3} \cmidrule(lr){4-10}
			& {F} & {M} & {B} & {EA} & {I} & {LH} & {ME} & {SA} & W \\
            \midrule
            {angry} & 6962 & \bfseries 17803 & 1804 & 680 & 540 & 2420 & 2072 & 267 & \bfseries 16982 \\
            {disgust} & 1733 & \bfseries 2054    & 285 & 173 & 61 & 334 & 328 & 48 & \bfseries 2558 \\
            {fear} & 3117 & \bfseries 3222    & 341 & 314 & 120 & 390 & 451 & 73 & \bfseries 4650 \\
            {happy} & \bfseries 79797 & 54241  & 8858 & 7917 & 2395 & 12038 & 6373 & 1935 & \bfseries 94522 \\
            {neutral} & 33708 & \bfseries 40858  & 6313 & 4644 & 1517 & 7109 & 6136 & 1016 & \bfseries 47831 \\
            {sad} & 11175 & \bfseries 14156  & 2002 & 1823 & 535 & 2262 & 2067 & 519 & \bfseries 16123 \\
            {surprise} & 6439 & \bfseries 7588    & 903 & 913 & 169 & 1017 & 1017 & 146 & \bfseries 9862 \\
            \midrule
            {Total} & \bfseries 142931 & 139922 & 20506 & 16464 & 5337 & 25570 & 18444 & 4004 & \bfseries 192528 \\
            \bottomrule
		\end{tabular}
	\end{table*}

    \subsection{Bias induction}\label{ssec:induction}

    Our approach to inducing bias has two main goals. First, we aim to push the bias to its most extreme cases while maintaining precise control over the specific type and strength of the induced bias. Second, we must ensure that the trained models remain comparable and perform similarly overall. To ensure comparability, we impose key constraints: the dataset size must remain constant, and the relative proportions of each emotion (target class) must stay consistent, regardless of the bias level. To achieve the first objective while adhering to these constraints, we implement the following bias induction methodology.

    \begin{table}[h]
        \sisetup{
            table-alignment-mode = format,
            table-number-alignment = right,
            table-text-alignment = right,
            table-format=5.0,
            group-minimum-digits = 4,
            detect-weight=math,
            mode=text
            }
		\centering
        \caption{Balanced baseline for race}\label{tab:cont_balanced_race}
		\begin{tabular}{lSSSSS}
			\toprule
            \multirow{2}{*}{\textbf{Class}} & \multicolumn{5}{c}{\textbf{Race}} \\
            \cmidrule(lr){2-6}
			& {B} & {EA} & {LH} & {ME} & {W} \\
            \midrule
            angry & 680 & 680 & 680 & 680 & 680 \\
            disgust & 173 & 173 & 173 & 173 & 173 \\
            fear & 314 & 314 & 314 & 314 & 314 \\
            happy & 6373 & 6373 & 6373 & 6373 & 6373 \\
            neutral & 4644 & 4644 & 4644 & 4644 & 4644 \\
            sad & 1823 & 1823 & 1823 & 1823 & 1823 \\
            surprise & 903 & 903 & 903 & 903 & 903 \\
            \midrule
            Total & 14910 & 14910 & 14910 & 14910 & 14910 \\
            \bottomrule
        \end{tabular}
	\end{table}

    \begin{table}[h]
        \sisetup{
            table-alignment-mode = format,
            table-number-alignment = right,
            table-text-alignment = right,
            table-format=5.0,
            group-minimum-digits = 4,
            detect-weight=math,
            mode=text
            }
    	\centering
        \caption{Example bias configuration. Overrepresented groups are in \textbf{bold}, and underrepresented ones are in \textit{italics}.}\label{tab:conf_bias_race}
        \begin{tabular}{lrrrrr}
            \toprule
            \multirow{2}{*}{\textbf{Class}} & \multicolumn{5}{c}{\textbf{Race}} \\
            \cmidrule(lr){2-6}
            & {B} & {EA} & {LH} & {ME} & {W} \\
            \midrule
            angry & \itshape 0.00 & \itshape 0.00 & \itshape 0.00 & \itshape 0.00 & \bfseries 1.00 \\
            disgust & 0.20 & 0.20 & 0.20 & 0.20 & 0.20 \\
            fear & 0.20 & 0.20 & 0.20 & 0.20 & 0.20 \\
            happy & 0.20 & 0.20 & 0.20 & 0.20 & 0.20 \\
            neutral & 0.20 & 0.20 & 0.20 & 0.20 & 0.20 \\
            sad & \bfseries 0.25 & \bfseries 0.25 & \bfseries 0.25 & \bfseries 0.25 & \itshape 0.00 \\
            surprise & 0.20 & 0.20 & 0.20 & 0.20 & 0.20 \\
            \bottomrule
        \end{tabular}
    \end{table}

    \begin{table}[h]
        \sisetup{
            table-alignment-mode = format,
            table-number-alignment = right,
            table-text-alignment = right,
            table-format=5.0,
            group-minimum-digits = 4,
            detect-weight=math,
            mode=text
            }
    	\centering
        \caption{Example biased dataset. Overrepresented groups are in \textbf{bold}, and underrepresented ones are in \textit{italics}.}\label{tab:cont_biased_race}
        \begin{tabular}{lrrrrr}
            \toprule
            \multirow{2}{*}{\textbf{Class}} & \multicolumn{5}{c}{\textbf{Race}} \\
            \cmidrule(lr){2-6}
            & {B} & {EA} & {LH} & {ME} & {W} \\
            \midrule
            angry & \itshape 0 & \itshape 0 & \itshape 0 & \itshape 0 & \bfseries 680 \\
            disgust & 34 & 34 & 34 & 34 & 34 \\
            fear & 62 & 62 & 62 & 62 & 62 \\
            happy & 1,274 & 1,274 & 1,274 & 1,274 & 1,274 \\
            neutral & 928 & 928 & 928 & 928 & 928 \\
            sad & \bfseries 455 & \bfseries 455 & \bfseries 455 & \bfseries 455 & \itshape 0 \\
            surprise & 180 & 180 & 180 & 180 & 180 \\
            \midrule
            Total & 2,933 & 2,933 & 2,933 & 2,933 & 3,158 \\
            \bottomrule
        \end{tabular}
    \end{table}
    
    \begin{enumerate}
        \item \textbf{Generation of a balanced baseline.} For a given demographic component $A$, we generate a balanced dataset by undersampling the original dataset $X$ such that for each class $y \in \mathcal{C}$, all demographic groups have equal representation. To do this, for each class $y$, we identify the minimum group size $n_y^{\text{min}} = \min_{a \in \mathcal{A}} n_y^a$ and subsample each group $a \in \mathcal{A}$ to contain exactly $n_y^{\text{min}}$ samples.

        Due to dataset size limitations, this balancing is performed independently for each demographic variable. For race, we restrict $\mathcal{A}$ to exclude Indian and Southeast Asian groups due to insufficient representation. While this results in different-sized balanced datasets for gender and race, both maintain sufficient samples for model training. Table~\ref{tab:cont_balanced_race} illustrates the outcome of this balancing process for the race component.

        \item \textbf{Bias induction.} Bias is induced by undersampling the balanced dataset, ensuring that all resulting datasets for a given demographic component are of the same size. We consider a proportion $p^a_y\in [0,1]$ of samples to be chosen from each class $y\in \mathcal{C}$ and demographic group $a \in \mathcal{A}$. We impose that the sum of proportions for each class equals one, i.e., $\sum_{a\in \mathcal{A}} p^a_y = 1$, ensuring that the resulting dataset will always have $\frac{n_y}{|\mathcal{A}|}$ samples from each class $y$, where $n_y$ refers to the number of examples of class $y$ in the balanced dataset. This method allows us to induce cases with extreme representational and stereotypical bias while maintaining the same dataset sizes. 
        
        An example bias configuration is provided in Table~\ref{tab:conf_bias_race}, with the resulting dataset represented in Table~\ref{tab:cont_biased_race}. This particular configuration simulates stereotypical bias where the White race group is overrepresented in the \textit{angry} class ($p^W_{\text{angry}}=1$) and underrepresented in the \textit{sad} class, while keeping the other emotions balanced ($p^W_{\text{sad}}=0$ and $p^a_{\text{sad}}=0.25$ for every $a\ne W$). The remaining emotions are left balanced ($p^a_y=0.2$ for all $y\notin \{\text{angry},\text{sad}\}$ and $a\in \mathcal{A}$).
    
    \end{enumerate}
    
    To facilitate the definition and analysis of results, it is essential that the different bias configurations include an unbiased reference point. However, the bias definition used so far, based on the proportion $p$ of one group, does not always produce reference points, or these occur at different values of $p$. For instance, for binary demographic variables (e.g., gender, $|\mathcal{A}|=2$), an unbiased situation means $p=0.5$, while for variables with more than two groups (e.g., race, $|\mathcal{A}|=5$), balance is represented by a different value ($p=0.2$ in the case of race).

    Therefore, we introduce an auxiliary bias parameter $f\in[-1,1]$, where $f=0$ always encodes absence of bias, $f=-1$ indicates maximum underrepresentation of the target group, and $f=1$ maximum overrepresentation. From this parameter $f$, we can obtain $p$ as:

    \begin{equation} p = \left(\frac{f+1}{2}\right)^{\log_2{|\mathcal{A}|}}. \label{eq:transform}\end{equation}

    Note that in all cases the resulting $p$ ranges from $0$ when $f=-1$ to $1$ when $f=1$. The reference point varies depending on $|\mathcal{A}|$, i.e., in the gender component $p=0.5$ for $f=0$, while on race $p=0.2$ for the same $f$.
    
    Based on this parameter $f$, we can simulate the two main types of biases:

    \begin{outline}
         \1 \textbf{Representational bias}. To simulate representational bias, we use $f^a$ to determine the global proportion $p^a$ of a specific group $a\in \mathcal{A}$ across all emotions. For this group, we set an equal proportion $p^a$ for all classes, that is, $p^a_y = p^a \text{ for all } y \in \mathcal{C}$. The remaining proportion in each class is distributed equally between the rest of the groups $a'$, setting $p^{a'}_y = (1 - p^a) / (|\mathcal{A}| - 1)$ for each of them. 
         \1 \textbf{Stereotypical bias}. In stereotypical bias, one or more classes $y \in \mathcal{C}$ exhibit different demographic composition than the others. To simulate this, we use different $f^a_y$ to define the proportion $p^a_y$ of a specific group $a\in A$ for each biased class, while the remaining proportion of those classes is distributed equally among the other groups $a'$, each assigned $p^{a'}_y = (1 - p^a_y) / (|\mathcal{A}| - 1)$. For the remaining classes, we keep balance among all the groups.
     \end{outline}

    \subsection{Overview of experiments}\label{ssec:overview}

    \begin{table*}[hbt]
        \centering
        \caption{Bias configuration for the experiments. The bias proportion $p$ is obtained after applying the transformation (Eq. (\ref{eq:transform})) to $f$. In the multiclass settings (c) and (f), $p_+$ is obtained by transforming $f$ while $p_-$ is obtained by transforming $-f$.
        For brevity, in (d), (e) and (f), some rows only show the formula used to calculate the proportion that is assigned to B, EA, LH and ME.
        }
        \label{tab:confs}
        \subfloat[Gender, Exp. 1: Representational bias.]{
            \begin{minipage}{0.294\linewidth}
            \centering
            \begin{tabular}{lrr}
                \toprule
                \multirow{2}{*}{\textbf{Class}} & \multicolumn{2}{c}{\textbf{Gender}} \\
                \cmidrule(lr){2-3}
                & M & F \\
                \midrule
                   angry & $1-p$ & $p$ \\
                 disgust & $1-p$ & $p$ \\
                    fear & $1-p$ & $p$ \\
                   happy & $1-p$ & $p$ \\
                 neutral & $1-p$ & $p$ \\
                     sad & $1-p$ & $p$ \\
                surprise & $1-p$ & $p$ \\
                \bottomrule
            \end{tabular}
            \end{minipage}
        }\quad%
        \subfloat[Gender, Exp. 2: Single class stereotypical bias.]{ 
            \begin{minipage}{0.31\linewidth}
            \centering 
            \begin{tabular}{lrr}
                \toprule
                \multirow{2}{*}{\textbf{Class}} & \multicolumn{2}{c}{\textbf{Gender}} \\
                \cmidrule(lr){2-3}
                & M & F \\
                \midrule
                angry & $0.5$ & $0.5$ \\
                disgust & $0.5$ & $0.5$ \\
                fear & $0.5$ & $0.5$ \\
                happy & $1-p$ & $p$ \\
                neutral & $0.5$ & $0.5$ \\
                sad & $0.5$ & $0.5$ \\
                surprise & $0.5$ & $0.5$ \\
                \bottomrule
            \end{tabular}
            \end{minipage}
        }\quad%
        \subfloat[Gender, Exp. 3: Multiclass stereotypical bias.]{
            \begin{minipage}{0.31\linewidth}
            \centering
            \begin{tabular}{lrr}
                \toprule
                \multirow{2}{*}{\textbf{Class}} & \multicolumn{2}{c}{\textbf{Gender}} \\
                \cmidrule(lr){2-3}
                & M & F \\
                \midrule
                angry & $1-p_+$ & $p_+$ \\
                disgust & $0.5$ & $0.5$ \\
                fear & $0.5$ & $0.5$ \\
                happy & $0.5$ & $0.5$ \\
                neutral & $0.5$ & $0.5$ \\
                sad & $1-p_-$ & $p_-$ \\
                surprise & $0.5$ & $0.5$ \\
                \bottomrule
            \end{tabular}
            \end{minipage}
        }

        \vspace{1em}
         
        \subfloat[Race, Exp. 1: Representational bias.]{
            \begin{minipage}{0.294\linewidth}
        	\centering
            \resizebox{\linewidth}{!}{
            \begin{tabular}{lrrrrr}
                \toprule
                \multirow{2}{*}{\textbf{Class}} & \multicolumn{5}{c}{\textbf{Race}} \\
                \cmidrule(lr){2-6}
                & B & EA & LH & ME & W \\
                \midrule
                   angry & \multicolumn{4}{c}{\dotfill$(1-p)/4$\dotfill} & $p$ \\
                 disgust & \multicolumn{4}{c}{\dotfill$(1-p)/4$\dotfill} & $p$ \\
                    fear & \multicolumn{4}{c}{\dotfill$(1-p)/4$\dotfill} & $p$ \\
                   happy & \multicolumn{4}{c}{\dotfill$(1-p)/4$\dotfill} & $p$ \\
                 neutral & \multicolumn{4}{c}{\dotfill$(1-p)/4$\dotfill} & $p$ \\
                     sad & \multicolumn{4}{c}{\dotfill$(1-p)/4$\dotfill} & $p$ \\
                surprise & \multicolumn{4}{c}{\dotfill$(1-p)/4$\dotfill} & $p$ \\
                \bottomrule
            \end{tabular}}
            \end{minipage}
        }\quad%
        \subfloat[Race, Exp. 2: Single class stereotypical bias.]{
            \begin{minipage}{0.31\linewidth}
        	\centering
            \resizebox{\linewidth}{!}{
            \begin{tabular}{lrrrrr}
                \toprule
                \multirow{2}{*}{\textbf{Class}} & \multicolumn{5}{c}{\textbf{Race}} \\
                \cmidrule(lr){2-6}
                & B & EA & LH & ME & W \\
                \midrule
                angry & $0.2$ & $0.2$ & $0.2$ & $0.2$ & $0.2$ \\
                disgust & $0.2$ & $0.2$ & $0.2$ & $0.2$ & $0.2$ \\
                fear & $0.2$ & $0.2$ & $0.2$ & $0.2$ & $0.2$ \\
                happy & \multicolumn{4}{c}{\dotfill$(1-p)/4$\dotfill} & $p$ \\
                neutral & $0.2$ & $0.2$ & $0.2$ & $0.2$ & $0.2$ \\
                sad & $0.2$ & $0.2$ & $0.2$ & $0.2$ & $0.2$ \\
                surprise & $0.2$ & $0.2$ & $0.2$ & $0.2$ & $0.2$ \\
                \bottomrule
            \end{tabular}}
            \end{minipage}
        }\quad%
        \subfloat[Race, Exp. 3: Multiclass stereotypical bias.]{
            \begin{minipage}{0.31\linewidth}
        	\centering
            \resizebox{\linewidth}{!}{
            \begin{tabular}{lrrrrr}
                \toprule
                \multirow{2}{*}{\textbf{Class}} & \multicolumn{5}{c}{\textbf{Race}} \\
                \cmidrule(lr){2-6}
                & B & EA & LH & ME & W \\
                \midrule
                angry & \multicolumn{4}{c}{\dotfill$(1-p_+)/4$\dotfill} & $p_+$ \\
                disgust & $0.2$ & $0.2$ & $0.2$ & $0.2$ & $0.2$ \\
                fear & $0.2$ & $0.2$ & $0.2$ & $0.2$ & $0.2$ \\
                happy & $0.2$ & $0.2$ & $0.2$ & $0.2$ & $0.2$ \\
                neutral & $0.2$ & $0.2$ & $0.2$ & $0.2$ & $0.2$ \\
                sad & \multicolumn{4}{c}{\dotfill$(1-p_-)/4$\dotfill} & $p_-$ \\
                surprise & $0.2$ & $0.2$ & $0.2$ & $0.2$ & $0.2$ \\
                \bottomrule
            \end{tabular}}
            \end{minipage}
        }
    \end{table*}

    To examine how bias propagates from datasets to models across various configurations, we design three experiments using the bias induction method described in Section~\ref{ssec:induction}. Specifically, for each bias configuration, we generate datasets that are biased in terms of their demographic components of gender and race. For gender, the induced bias definition is symmetric, as $|\mathcal{A}|=2$, so we arbitrarily choose Female as the target demographic group. For race, we select White as the target group, aligning with the typical representational bias observed in FER datasets, where the White group is often significantly overrepresented.

    \begin{itemize}
        \item \textit{Experiment 1. Representational bias.} For this experiment, we simulate a bias scenario where only representational bias is induced. This situation is similar to that of FER datasets created under controlled conditions, where at least one sample of each emotion is taken for each subject, avoiding stereotypical bias. The bias configurations for gender and race can be observed in parts (a) and (d) of Table~\ref{tab:confs}, respectively.
    
        \item \textit{Experiment 2. Single class stereotypical bias.} For this experiment, we simulate a stereotypical bias situation where a single class is biased, with the rest of the dataset balanced. We choose the class \textit{happy} to induce stereotypical bias by varying $f^{\text{happy}}_a$ for the target group of each demographic variable, and leaving the rest of the classes balanced ($f^y_a=0$ for $y\neq \text{happy}$). We consider the \textit{happy} class as it is the most stereotypically biased one in many in-the-wild datasets~\cite{Dominguez-Catena2024}. The bias configuration tables can be seen in parts (b) for gender and (e) for race of Table~\ref{tab:confs}.

        \item \textit{Experiment 3. Multiclass stereotypical bias.} For this experiment, we simulate a more complex bias scenario, with two classes biased in opposite ways. We choose two similarly sized classes, \textit{sad} and \textit{angry}, to reduce the side effect of representational bias induction. We use $f^a_{\text{angry}}$ to modify the proportion $p^+$ of the target group $a\in A$ for class \textit{angry}, and use $f^a_{\text{sad}} = -f^a_{\text{angry}}$ to induce the opposite modification for class \textit{sad}, resulting in a proportion $p^-$. Again, we distribute the remaining proportions equally among the other groups in those classes.

        A summary of these bias configurations can be seen in parts (c) and (f) of Table~\ref{tab:confs}, for gender and race, respectively.
    \end{itemize}

    For each configuration, $f$ ranges from $-1$ to $1$ in increments of $0.2$, for both gender and race demographic components. Each experiment generates a total of $22$ datasets ($11$ for all increments of $0.2$ in the range $[0,1]$, for each demographic component), leading to $66$ different datasets used to train $66$ potentially biased models. Additionally, each experiment is run three times with different random seeds, resulting in a total of $198$ pairs of datasets and models.

\subsection{Experimental setup}\label{ssec:setup}

    We train a ResNet50 network~\cite{He2015} on each biased dataset, starting with the pre-trained ImageNet-1K weights from PyTorch. ResNet50 is a standard architecture for image classification, and the PyTorch weights provide a common baseline. Training is done under a \textit{1cycle} policy~\cite{Smith2018} with a maximum learning rate of $10^{-4}$ for $20$ epochs.

    The trained models are evaluated on the full AffectNet test set. These predictions are then used to compute the model bias metrics presented in Section~\ref{ssec:measurement_model}.
    
    To mitigate randomness from subsampling and training, we repeat the dataset generation, training, and evaluation three times per configuration. We then use the average of the three repetitions for each reported metric.
    
    The experiments are coded in PyTorch 2.0.1 on a PC with a GeForce RTX 3060 Super GPU, 20~GB RAM, and an Intel\textregistered\ Xeon\textregistered\ i5-8500 CPU, running Ubuntu Linux 22.04.

\section{Results}\label{sec:results}

    In this section, we analyze the results of the conducted experiments. Our main objectives are to identify how bias in datasets propagates to bias in models, and to validate our proposed adaptations of model fairness constraints as bias metrics. To achieve these goals, we decompose the analysis into three phases.

    \begin{outline}[enumerate]
        \1 \textbf{Dataset bias}. First, in Section~\ref{ssec:res_dataset_bias}, we validate the bias induction process using the dataset bias metrics introduced in Section~\ref{ssec:measurement_dataset}.
        \1 \textbf{Model bias}. Next, Section~\ref{ssec:res_model_bias} focuses on the bias observed in the trained models, examining the behavior of the various model bias metrics discussed in Section~\ref{ssec:measurement_model}, including the newly proposed model bias metrics.
        \1 \textbf{Bias correlation}. Finally, Section~\ref{ssec:res_bias_correlation} concludes the analysis with a study of the correlation between dataset and model bias metrics. This study enables us to identify which notions of fairness are infringed upon when certain types of dataset bias are induced.
    \end{outline}
    
    \subsection{Measured dataset bias}\label{ssec:res_dataset_bias}

\begin{figure}[htb]
        \centering
		\includegraphics[width=\columnwidth]{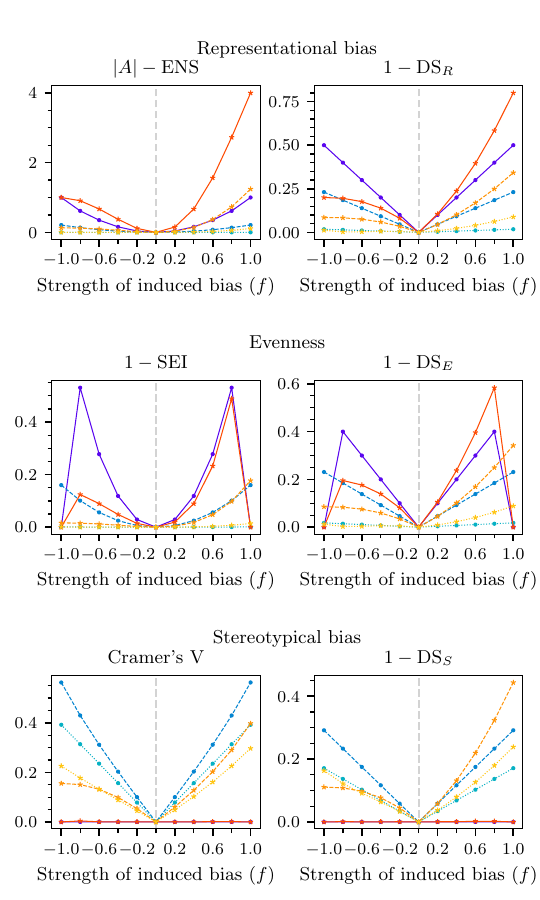}
        \includegraphics[width=0.65\columnwidth]{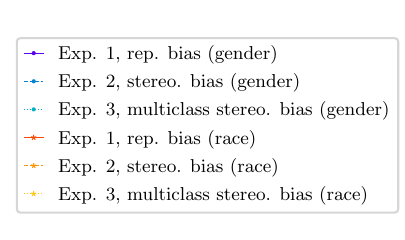}
		\caption{Measured dataset bias according to different metrics.}
		\label{fig:measured_dataset_bias}
	\end{figure}

    The dataset bias measured after the induction process is shown in Figure~\ref{fig:measured_dataset_bias}. This figure presents the bias using six different metrics: three classical metrics (general representational bias, evenness, and stereotypical bias) and three DSAP-based metrics. The induced bias parameter $f$ is reflected in the horizontal axis, with the unbiased reference at $f=0$.

    \vspace{0.5em}\noindent\textbf{Representational bias} metrics (ENS and $\text{DS}_R$) respond as expected to induced representational bias, showing the unbiased baseline at $f=0$ with bias increasing towards the extremes. Both metrics exhibit similar trends but with different response curves, with $\text{DS}_R$ showing a more linear response to $f$. The experiments focused on stereotypical bias also show representational bias as a side effect of the induction process, though weaker than the experiments focused on representational experiments on the same demographic variables. This residual bias is more prominent in the single class scenario, whereas the multiclass scenario mitigates this by affecting the two classes in opposite ways. In the racial bias experiments, the induced bias is greater for $f>0$ than for $f<0$, reflecting the natural asymmetry when $|\mathcal{A}|>2$: extreme positive bias ($f=1$) results in a single group's presence, while extreme negative bias ($f=-1$) maintains four different groups.

    \vspace{0.5em}\noindent\textbf{Evenness}, as a component of representational bias, is measured by SEI and $\text{DS}_E$. These metrics show similar trends to general representational bias metrics, with $\text{DS}_E$ particularly close to $\text{DS}_R$. The key difference appears at $f=\pm1$, where abrupt changes in evenness are registered as groups appear or disappear from the dataset. SEI's robustness to the number of groups makes its results comparable between race and gender components for $f\in[0,1]$, though this comparability breaks down for $f<0$ due to the aforementioned asymmetry.

    \vspace{0.5em}\noindent\textbf{Stereotypical bias} metrics (Cramer's V and $\text{DS}_S$) show no variation in representationally biased experiments, while correctly capturing bias in stereotypically biased experiments. Both metrics demonstrate similar behavior, ranking single class stereotypical bias experiments as more biased than multiclass ones of the same demographic component. The key difference lies in their treatment of demographic components: $\text{DS}_S$ assigns higher relative importance to race configurations, while Cramer's V indicates greater bias in gender-based configurations.

    \subsection{Measured model bias}\label{ssec:res_model_bias}

\begin{figure*}[htb]
        \centering
		\includegraphics[]{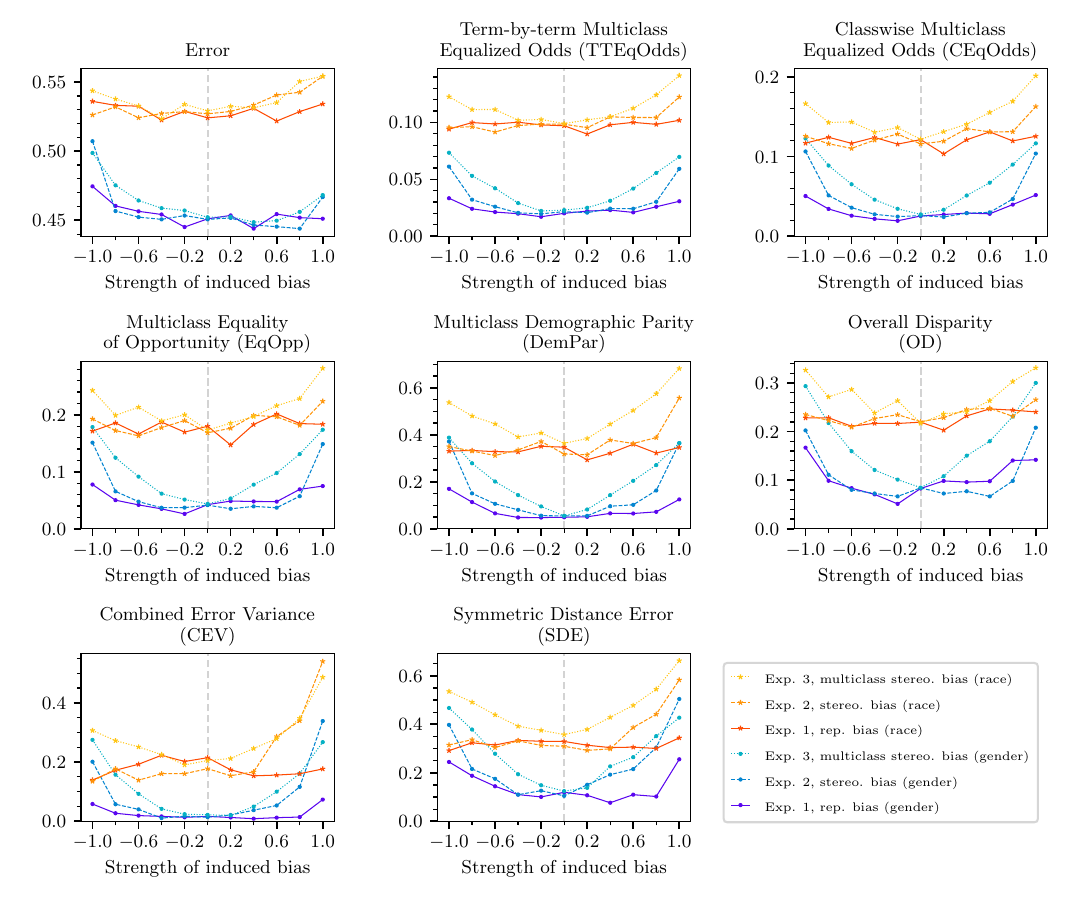}
  
		\caption{Measured model bias according to different metrics.}
		\label{fig:measured_model_bias}
	\end{figure*}

    The bias exhibited by the models trained on the biased datasets is shown in Figure~\ref{fig:measured_model_bias}. Each subplot shows the results according to a different metric from Section~\ref{ssec:measurement_model}. For all figures, the horizontal axis corresponds to the induced bias parameter $f$.

    \vspace{0.5em}\noindent\textbf{Model error:} The error rate, included as a performance reference, shows an $8\%$ increase in racially biased datasets compared to gender-biased ones, reflecting the impact of smaller training sets. Additionally, error rates increase towards extreme bias situations ($f=\pm1$), particularly in stereotypical bias experiments, reaching up to $5\%$ increase from $f=0$ to $f=-1$ in both single and multiclass experiments on the gender component. Notably, the impact in the gender component seems larger for negative biases (removing female presenting people) compared to positive ones.

    \vspace{0.5em}\noindent\textbf{Gender:} Most metrics show strongest bias propagation in the multiclass stereotypical experiment, followed by single class, and finally representational experiments. The multiclass experiment shows a more linear increase in propagated bias with respect to $f$, while the single class exhibits a slower initial response followed by a sharp increase near the extremes. In some metrics this sudden increase makes single class bias surpass multiclass bias at the extremes, such as DemPar, CEV and SDE at $f=1$. Around $f=0$, metrics show varying stability in propagated representational bias: while DemPar and CVE show little variation for $f\in[-0.6,0.6]$, others like CEqOdds, EqOpp and OD show significant changes, particularly a bias decrease at $f=-0.2$. Notably, CVE and DemPar come closer to 0 relative to their range at $f=0$ compared with the others. Their nearly flat response to representational bias contrasted with sharp increases for stereotypical bias as $|f|$ grows suggests they may be particularly suitable for differentiating between these types of bias.

    \vspace{0.5em}\noindent\textbf{Race:} With five represented groups instead of two, metrics generally show higher baseline bias scores and weaker response to induced bias compared to the gender component. TTEqOdds shows the largest gap between race and gender settings, behaving more similarly to the error rate and showing less sensitivity to induced bias. In contrast, DemPar, CVE and SDE show more overlap between gender and race components and a more stable response to variations in $f$. As in gender, most metrics indicate greatest bias propagation in multiclass settings, followed by single class and representational bias.

    Following the characteristics of the measured dataset bias, the race component exhibits strong asymmetry in single class experiments and, to a lesser extent, in multiclass experiments. This is particularly evident in DemPar, CEV and SDE, showing almost no bias variation for $f<0$ but doubling measured bias at $f=1$. In multiclass experiments, the asymmetry is reduced but still present, with most metrics indicating higher model bias at $f=1$. For representational bias, propagation appears relatively linear with $f$, except for CVE, which shows a counterintuitive decrease in bias as $|f|$ grows.

    \subsection{Bias metrics correlation}\label{ssec:res_bias_correlation}

\begin{figure}[htb]
        \centering
		\includegraphics[width=\linewidth]{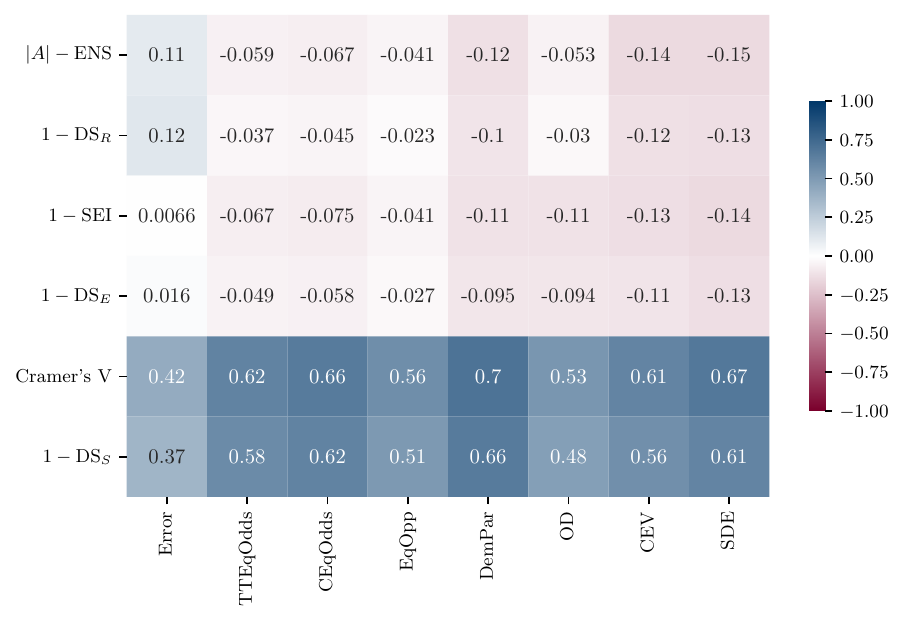}
  
		\caption{Spearman's $\rho$ rank correlation between the dataset and model bias metrics.}
		\label{fig:corr_dataset_to_model_bias}
	\end{figure}

    Figure~\ref{fig:corr_dataset_to_model_bias} shows the Spearman rank correlation ($\rho$) between dataset and model bias metrics. Note that we measure correlation, not the strength of the response of each metric. The correlation has been calculated independently for the race and gender components, and then averaged. We analyze the results from stronger to weaker propagation, that is, we start with stereotypical bias and then analyze representational bias, first in general and then focusing on evenness.

    \vspace{0.5em}\noindent\textbf{Stereotypical bias:} Stronger correlations are observed for stereotypical dataset bias metrics compared to representational and evenness ones. All model bias metrics show moderate correlations ($\rho>0.53$) with Cramer's V, followed closely by $\text{DS}_S$ ($\rho>0.48$). DemPar shows the strongest correlation overall ($\rho=0.7$), followed by SDE and CEqOdds ($\rho=0.67$ and $\rho=0.66$ respectively).

\begin{figure}[htb]
        \centering
		\includegraphics[width=\linewidth]{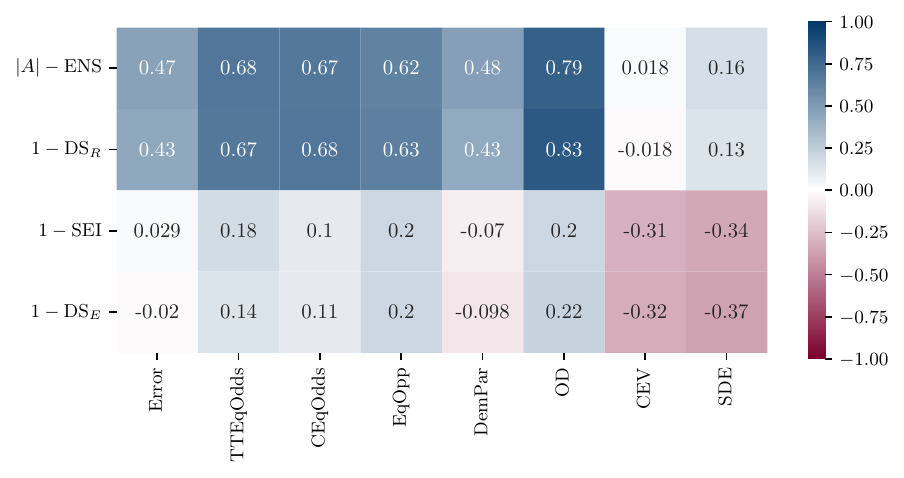}
  
		\caption{Spearman's $\rho$ rank correlation between the representational and evenness dataset bias metrics and all model bias metrics, restricted to only the results of the representational bias experiment.}
		\label{fig:corr_representational_bias}
	\end{figure}

    \vspace{0.5em}\noindent\textbf{Representational bias:} No significant correlations are found in the full experiment set, with all metrics showing correlations very weak negative correlations ($|\rho|<0.15$). To isolate the effect of representational bias from the stronger stereotypical bias effect, Figure~\ref{fig:corr_representational_bias} shows correlations restricted to the representational bias experiment. In these results, CEV and SDE fall dramatically and show a very weak correlation with representational bias, and negative with evenness. The rest show stronger correlations, with OD standing out with a very strong correlation ($\rho=0.83$) with $\text{DS}_R$, followed by both EqOdds variants ($\rho>0.67$). The correlations with evenness metrics are still weak or very weak in all cases.

\begin{figure}[htb]
        \centering
		\includegraphics[width=\linewidth]{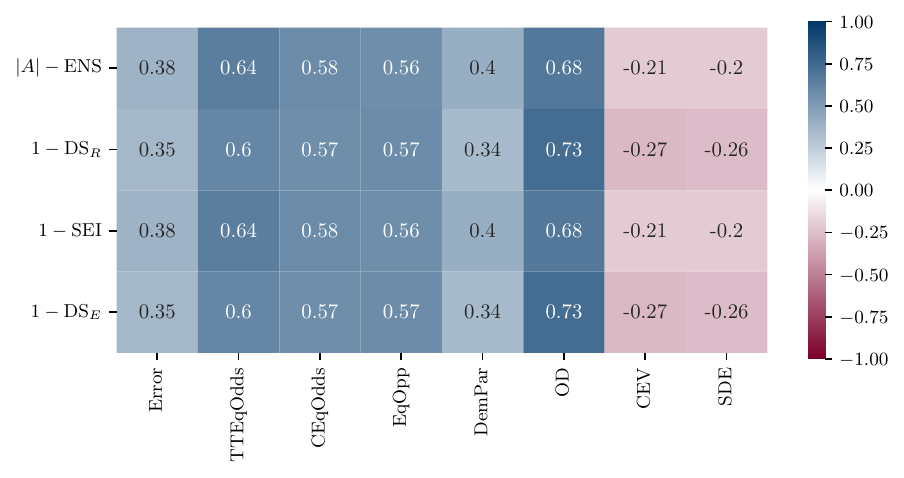}
  
		\caption{Spearman's $\rho$ rank correlation between the representational and evenness dataset bias metrics and all model bias metrics, restricted to only the results of the representational bias experiment, and excluding the $f=\pm1$ configurations.}
		\label{fig:corr_evenness}
	\end{figure}

    The low correlations for \textbf{evenness} in Figures~\ref{fig:corr_dataset_to_model_bias} and~\ref{fig:corr_representational_bias} are likely due to sudden changes in measured evenness when a group disappears ($f=\pm1$), which is uncommon in real scenarios. Figure~\ref{fig:corr_evenness} shows correlations for the representational bias experiment excluding these extreme cases. By removing the measurements at $f=\pm1$, the evenness metrics become almost identical to the general representational bias metrics. The results, again, show a stronger correlation with OD ($\rho>0.68$), followed by the rest of the new adaptations ($\rho>0.34$), and CEV and SDE showing weak negative correlations.

\section{Discussion}\label{sec:discussion}

    Although previous results on the incompatibility of different model fairness metrics~\cite{Miconi2017} indicate that the possible notions of fairness are radically different, our results suggest that these differences are less clear when dealing with unfair models. We observe \textbf{strong agreement} between our chosen set of multiclass bias metrics, even when forcing extreme dataset biases in six different scenarios. The four proposed adaptations of classical fairness constraints, as well as the preexisting model bias metrics, demonstrate strong correlations with dataset bias metrics and similar behaviors overall.

    The analysis of correlations between dataset and model bias metrics reveals several key findings:

    \begin{itemize}
        \item Model bias metrics show stronger responses to stereotypical bias compared to representational bias in the training data.

        \item This difference aligns with previous research~\cite{Dominguez-Catena2024c} showing weaker effects of representational bias in FER, as evidenced by its lower impact on model error rates.

        \item In the absence of stereotypical bias, most metrics maintain strong correlations with representational bias. Among all metrics, OD exhibits the strongest response to representational bias, while CEV and SDE drop to very weak correlations.

        \item All metrics effectively identify the effects of stereotypical bias, with DemPar showing the strongest correlations. This metric exhibits more stable responses to the induced bias factor $f$ even in multiclass settings.
    \end{itemize}

    The strong correlations for OD and DemPar do not necessarily imply that these bias metrics are inherently better than the others, but rather that they accurately measure the propagated bias for our induced bias types. The fact that different model bias metrics correlate best with representational and stereotypical dataset bias, including the change in correlation direction for some metrics (CEV and SDE) supports the notion that different types of dataset bias distinctly affect model bias. Unfortunately, the current set of metrics does not completely disentangle these different types of dataset bias, lacking metrics that focus on measuring the effects of each type independently.

    Observing the magnitude of the measured model bias in our experiments, we confirm that in FER, representational bias tends to propagate more weakly into the model than stereotypical bias. Additionally, we observe that stereotypical bias propagates more strongly when it affects multiple classes at the same time, with current dataset bias metrics failing to account for this fact.

    Our experiments reveal an unexpected relationship between model fairness and accuracy: induced dataset bias leads to both increased model bias and decreased model performance, evidenced by strong correlations between error rates and bias metrics, especially for stereotypical bias. This finding challenges the commonly accepted \textbf{fairness-accuracy trade-off}~\cite{Menon2018}. Instead, in line with recent studies~\cite{Cooper2021,Gardner2022}, our results suggest that in FER, models trained on biased datasets tend to be both less fair and less accurate. This effect is observed when evaluating against a common test set whose naturally occurring biases, while present, are significantly lower than our artificially induced training biases. This controlled evaluation setting suggests that, at least for dataset-induced bias in FER, fairness and accuracy might be complementary rather than competing objectives. Further research is needed to fully validate this hypothesis.

    \subsection{Implications for FER}

    Our findings have several important implications for the development and deployment of FER systems:
    
    \begin{itemize}
    \item \textbf{Dataset Design:} While achieving perfect demographic balance may not be crucial due to the weaker propagation of representational bias, special attention should be paid to avoiding emotion-specific demographic patterns, as stereotypical bias shows stronger propagation effects. This is particularly crucial when multiple emotions are affected, as we observe compounding effects when stereotypical bias is not constrained to a single class. Furthermore, this effect is not reflected in current dataset bias metrics.
    
    \item \textbf{Model Evaluation:} The strong correlation between different bias metrics in unfair scenarios suggests that a full analysis may not be necessary for initial bias screening. We recommend DemPar for stereotypical bias analysis and OD for representational bias, noting that OD performs well only in the absence of stereotypical bias. Nevertheless, the use of summarizing metrics that condense bias into single numerical values, particularly for complex problems like FER, should not be an excuse to avoid manual analysis of the effects of bias on model predictions.
    
    \item \textbf{FER Implementation:} Our results challenge the notion that fairness in FER comes at the cost of accuracy. This suggests that efforts to reduce dataset bias could improve both the fairness and the overall performance of FER systems, particularly in real-world applications where test conditions may differ significantly from biased training data.
    
    \end{itemize}

    \subsection{Limitations}
    Finally, there are some limitations to this work that should be addressed in future research. First, we base our results on predicted demographic properties, following on previous research~\cite{Yang2022,Ghosh2022}. This demographic labels are, unfortunately, not guaranteed to be exact, and limit the precision of our results. Additionally, although our choice of demographic components codified both binarily and with multiple groups covers more cases than previous research, these results could vary for ordinal demographic variables, such as age, or for variable codifications that support multiple labels simultaneously. Despite these shortcomings, our results already highlight some new behaviors of model bias when understood under the influence of properly measured dataset bias.

\section{Conclusion and Future Work}\label{sec:conclusion}

    This paper investigates how demographic bias propagates from datasets to models in Facial Expression Recognition (FER), a multiclass problem where bias involves multiple demographic variables, each potentially encoded with multiple groups (e.g., binary gender, multi-group race). We developed a methodology to systematically study this propagation by inducing controlled amounts of representational and stereotypical bias in the AffectNet dataset, then measuring their effects on trained models using both existing and newly proposed bias metrics.

    Our analysis reveals distinct propagation patterns for different types of bias. Stereotypical bias, particularly when affecting multiple emotion classes simultaneously, shows stronger propagation into model predictions compared to representational bias. This suggests that in FER dataset design, avoiding emotion-specific demographic patterns should take priority over achieving perfect demographic balance. Furthermore, our results challenge the common fairness-accuracy trade-off assumption: models trained on biased datasets tend to perform worse overall, indicating that bias mitigation could improve both fairness and accuracy.

    We find that while current model bias metrics strongly correlate when measuring unfair models, they show varying sensitivity to different types of dataset bias, suggesting that representational and stereotypical bias affect models in fundamentally different ways. We observe a stronger response to stereotypical bias, which could reflect either a weaker propagation of representational bias to model behavior in FER, or limitations in current metrics that may obscure important representational effects. Our proposed adaptations of classical fairness metrics to multiclass and multi-group scenarios perform well, showing more robust correlations with induced bias than previous metrics.

    Future work should address the development of metrics capable of independently measuring the effects of different types of bias propagation, particularly for cases where stereotypical bias affects multiple classes. Additionally, new model bias metrics are needed to better distinguish between the impacts of representational and stereotypical bias, along with specialized mitigation strategies for multiclass stereotypical scenarios. Our methodology could also be extended to study how other variables, such as dataset properties and model architectures, influence bias propagation in FER systems.

\section{Funding}

    This work was funded by a predoctoral fellowship and open access funding from the Research Service of the Universidad Publica de Navarra, the Spanish MICIN (PID2022-136627NB-I00/AEI/10.13039/501100011033 FEDER, UE), the Government of Navarre (0011-1411-2020-000079 -- Emotional Films), and the support of the 2024 Leonardo Grant for Researchers and Cultural Creators from the BBVA Foundation. It was also supported by the Flemish Government under the “Onderzoeksprogramma Artificiele Intelligentie (AI) Vlaanderen” programme, and the European Union (ERC, VIGILIA, 101142229).
    The BBVA Foundation, the European Union, and the European Research Council Executive Agency are not responsible for the opinions, comments, and content included in the project and/or its derived results, which are the sole responsibility of the authors.
    
\FloatBarrier
\bibliographystyle{IEEEtran}
\bibliography{fulllibrary}

\begin{IEEEbiography}[{\includegraphics[width=1in,height=1.25in,clip,keepaspectratio]{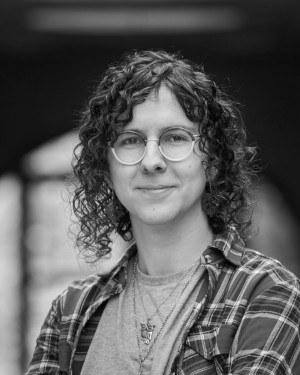}}]{Iris Dominguez-Catena} 
received the M.Sc. and Ph.D. degrees in computer science from the Public University of Navarra, in 2020 and 2024, respectively. He is currently Assistant Professor with the Public University of Navarra, researching on demographic bias issues in Artificial Intelligence. His research interests focus on AI fairness, bias detection and mitigation, and other ethical problems of AI deployment in society.
\end{IEEEbiography}

\begin{IEEEbiography}[{\includegraphics[width=1in,height=1.25in, clip, keepaspectratio]{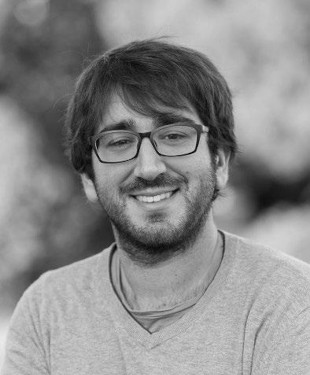}}]{Daniel Paternain}
received the M.Sc. and Ph.D. degrees in computer science from the Public University of Navarra, Pamplona, Spain, in 2008 and 2013, respectively. He is currently Associate Professor with the Department of Statistics, Computer Science and Mathematics. He is also the author or coauthor of almost 40 articles in journals from JCR and more than 50 international conference communications. His research interests include both theoretical and applied aspects of information fusion, computer vision and machine learning.
\end{IEEEbiography}

\begin{IEEEbiography}[{\includegraphics[width=1in,height=1.25in,clip,keepaspectratio]{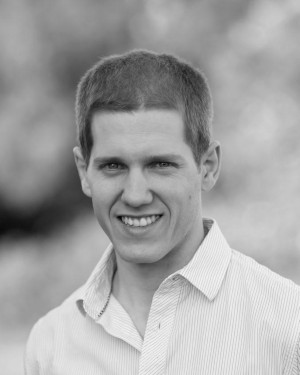}}]{Mikel Galar} (M14)
received the M.Sc. and Ph.D. degrees in computer science from the Public University of Navarra, Pamplona, Spain, in 2009 and 2012, respectively. He is currently Full Professor at the Public University of Navarra. He is the author of 54 published original articles in international journals and more than 90 contributions to conferences. He is a co-author of a book on imbalanced datasets and a book on large-scale data analytics. His research interests are  machine learning, deep learning, ensemble learning and big data. He  received the extraordinary prize for his PhD thesis from the Public University of Navarre and the 2013 IEEE Transactions on Fuzzy System Outstanding Paper Award (bestowed in 2016). 
\end{IEEEbiography}

\begin{IEEEbiography}[{\includegraphics[width=1in,height=1.25in,clip,keepaspectratio]{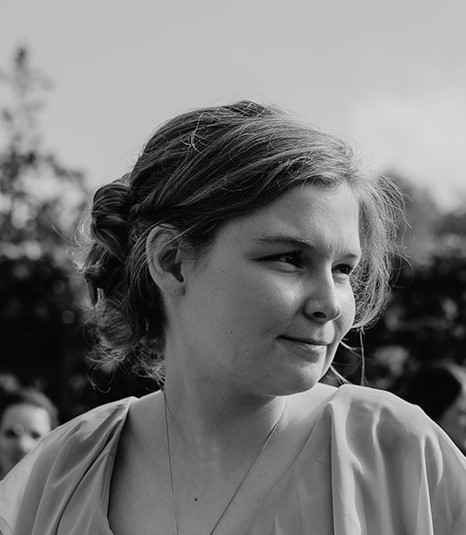}}]{MaryBeth Defrance}
received her M.Sc. in computer science engineering from Ghent University in 2022. She is currently in the third year of her Ph.D. in computer science engineering at Ghent University. She focuses on fairness in AI and more specifically on how fairness is measured. She has published in top conferences on evaluating bias mitigation methods and compatibility of fairness notions. 
\end{IEEEbiography}

\begin{IEEEbiography}[{\includegraphics[width=1in,height=1.25in,clip,keepaspectratio]{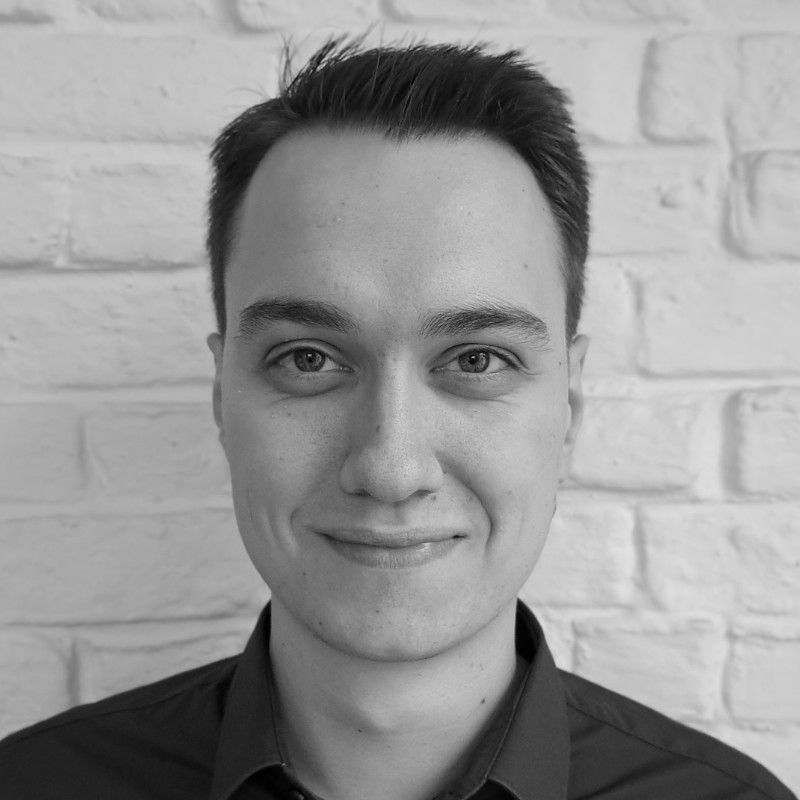}}]{Maarten Buyl}
\textbf{Maarten Buyl} received his M.Sc. and Ph.D. in computer science engineering from Ghent University in 2019 and 2023 respectively. He is currently a postdoctoral researcher at Ghent University and has been a Applied Science intern at Amazon in 2023 and a Visiting Postdoctoral Fellow at Harvard University in 2024. His work on AI fairness has been published in the top venues in the field, covering both the development of technical tools and the study of their limitations. 
\end{IEEEbiography}

\begin{IEEEbiography}[{\includegraphics[width=1in,height=1.25in,clip,keepaspectratio]{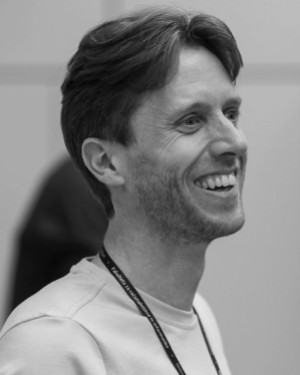}}]{Tijl De Bie}
Tijl De Bie received the Ph.D. degree in machine learning from KU Leuven, in 2005. He is currently a Senior Full Professor with Ghent University, Belgium. Before moving to Ghent University, he was a Reader with the University of Bristol, a Postdoctoral Researcher with KU Leuven and the University of Southampton, and a Visiting Research Scholar with UC Berkeley and UC Davis. He has previously held an ERC Consolidator Grant on the formalization of subjective interestingness in exploratory data mining. Currently he holds an ERC Advanced Grant on an innovative AI-based approach to mitigating the risks of mis-, des-, and malinformation. His current research interests include graph-based machine learning, recommender systems, data visualization, ethics and regulation of AI including the influence of AI on information integrity, and applications of AI, particularly in human resources and job market use cases.
\end{IEEEbiography}

\vfill

\end{document}